%% file: main.tex
\documentclass{article} % For LaTeX2e
\usepackage[T1]{fontenc}
\usepackage{iclr2027_conference,times}
\usepackage{fvextra}
\DefineVerbatimEnvironment{PromptBlock}{Verbatim}{
  breaklines=true,
  breakanywhere=true,
  fontsize=\footnotesize
}

\input{math_commands.tex}

\usepackage[table]{xcolor}
\definecolor{myBlue}{HTML}{0093B8}
\usepackage{amsmath,amssymb,amsfonts}
\usepackage{booktabs}
\usepackage{graphicx}
\usepackage{microtype}
\usepackage{natbib}
\usepackage{xcolor}
\usepackage{url}
\usepackage{enumitem}
\usepackage{algorithm}
\usepackage{algpseudocode}
\usepackage{multirow}
\usepackage{caption}
\usepackage{subcaption}
\usepackage{placeins}
\usepackage{hyperref}
\usepackage{wrapfig}
\usepackage{needspace}
\usepackage[most]{tcolorbox}
\usepackage{xcolor}

\definecolor{takeawayGray}{HTML}{555555}
\definecolor{takeawayGrayBack}{HTML}{F2F2F2}
\definecolor{comparisonGreen}{HTML}{19733B}
\definecolor{comparisonRed}{HTML}{B3261E}

\newtcolorbox{customblocktakeaway}{
  colframe=takeawayGray,
  colback=takeawayGrayBack,
  boxrule=0pt,
  left=4pt,
  right=5pt,
  top=5pt,
  bottom=3pt,
  arc=0pt,
  breakable,
  before skip=1.2\baselineskip,
  after skip=0.7\baselineskip,
  left skip=0pt,
  right skip=0pt,
  enhanced jigsaw,
  frame hidden,
  overlay={
    \draw[takeawayGray, line width=2pt]
      ([yshift=1pt]frame.north west) -- (frame.south west);
  },
  fontupper=\fontfamily{ptm}\selectfont,
  boxsep=1pt,
}

\title{Amortizing Sequential Experimental Design in Reasoning LLMs}

\author{Anita Kriz\thanks{Authors contributed equally.},\hspace{0.1em}
Antonin Berthon\footnotemark[1],\hspace{0.1em}
Nicolás Astorga,\hspace{0.1em}
Mihaela van der Schaar \\
University of Cambridge \\
Cambridge, UK \\
\texttt{\{ak2680,armb3\}@cam.ac.uk}
}

\iclrfinalcopy 
\begin{document}

\maketitle
\begin{abstract}
Despite their extensive pretraining knowledge and strong reasoning capabilities, large language models (LLMs) struggle to adaptively gather information in experimental design tasks. We study how to \emph{amortize sequential Bayesian experimental design (BED) into an LLM} by explicitly training the reasoning that guides action selection using information-gain rewards.
Our approach specifies the experimental generative model independently of the LLM's own beliefs, 
allowing us to optimize expected information gain under a model of our choosing. We investigate the design choices for amortization, including alternative generative-model constructions, reward estimation, hypothesis-space specification, multi-step objectives, and computational costs relative to training- and inference-time BED baselines.
On Twenty Questions, the resulting policies improve information acquisition and task performance, outperform inference- and training-time BED baselines at substantially lower computational cost, 
and generalize to unseen categories. We further demonstrate positive transfer from Twenty Questions training to Medi-Q, a real-world clinical dialogue benchmark.
These results suggest that information gathering can be learned as a generalizable reasoning skill in LLMs.

\end{abstract}

\section{Introduction}
Many real-world problems are not fully specified upfront, yet humans are remarkably effective at gathering the information needed to solve them~\citep{li2024mediq,huang2024mlagentbench}. Consider a clinician diagnosing a patient. 
They draw on their medical knowledge to ask about relevant symptoms, think about possible diagnoses, and identify tests that can distinguish between them. As new evidence is acquired, they refine their hypotheses to decide what to investigate next. Effective information gathering thus requires both the prior knowledge to inform decision-making and the ability to continuously adapt those decisions in light of new evidence.

Given their extensive pretraining knowledge and strong reasoning capabilities, large language models (LLMs) are increasingly being explored as agents in adaptive decision-making ~\citep{zhou2025large,ashokkumar2026large}. Despite this potential, LLMs have been shown to perform poorly in multi-turn settings~\citep{krishnamurthy2024can, zhou2025from,li2026questbench, laban2026llms}. In particular, \citet{gupta2025llmsbayesianoptimizationscientific} demonstrate that LLMs are insensitive to experimental feedback, suggesting that their performance relies more on their (potentially inaccurate) prior knowledge than on their ability to adapt their reasoning to new evidence.

In this work, \textit{we investigate how to improve LLMs' ability to adaptively gather information}. 
Rather than using an LLM to sample informative actions~\citep{kobalczyk2025active,choudhury2025bedllm,hartmann2026amortising}, 
we hypothesize that learning generalizable information gathering strategies requires explicitly training 
the reasoning process that guides action selection.
We formalize this problem through sequential Bayesian experimental design (BED)~\cite{rainforth2023modernbayesianexperimentaldesign}, 
which uses a generative model of the experiment to select actions that reduce uncertainty about an unknown quantity of interest. We investigate how to amortize this information-theoretic objective into an LLM reasoning policy using reinforcement learning with verifiable rewards (RLVR). Crucially, training-time amortization allows us to specify the experimental generative model \textit{independently} of the LLM's own beliefs. This means that we can evaluate actions based on their expected information gain under a model of our choosing.
We additionally investigate whether optimizing information gain over multiple interactions encourages non-myopic information-gathering strategies.
 
We empirically show that amortizing BED into a reasoning LLM develops strategies that generalize to previously unseen domains and tasks. Specifically, we train reasoning LLMs on Twenty Questions and demonstrate improved information acquisition across both seen and unseen categories. We show that models trained in this environment exhibit positive transfer to the real-world clinical dialogue benchmark Medi-Q~\citep{li2024mediq}. Our results show that decoupling the experimental generative model from the LLM's beliefs improves information acquisition.
We further find that rewarding expected information-gain rewards under our specified generative model improves learning relative to realized rewards, and that optimizing information acquisition over future interactions can improve sequential information gathering in non-myopic environments.

\textbf{Contributions.} Our contributions are:  
\begin{itemize}[label=$\blacktriangleright$]
\item We formulate amortized sequential BED for reasoning LLMs and study how to construct its generative model and information-gain reward (Sec.~\ref{sec:method}).
\item We demonstrate that information-gathering behavior can be amortized into a reasoning LLM, improving information acquisition and transferring across unseen domains and to a different information-gathering task (Sec.~\ref{sec:experiments}). 
\item We systematically study the design of amortized BED, including alternative generative-model constructions, reward estimation, hypothesis-space specification, multi-step objectives, and their computational cost relative to training- and inference-time BED baselines (Sec.~\ref{sec:experiments}).
\end{itemize}

\section{Background \& Problem Setup}

\subsection{Bayesian Experimental Design}
\label{sec:background_bed}
Let $\theta$ denote an unknown variable that we wish to learn about. To gain information about $\theta$, we can select an experimental design $x$, which results in an observation $y$. For example, in a diagnostic setting, $\theta$ may represent a patient's underlying condition that we want to identify, $x$ a diagnostic test, and $y$ the resulting test outcome. Because experiments may be subject to constraints such as cost or time, the chosen design $x$ should be as informative as possible. 

BED provides a framework for quantifying how informative a potential design $x$ is using a generative model of the experiment, $p(\theta,y\mid x) = p(\theta)p(y\mid\theta,x)$. Here, the prior $p(\theta)$ represents our initial beliefs about $\theta$ and the likelihood $p(y\mid\theta,x)$ describes the probability of observing $y$ given $\theta$ and $x$. After selecting $x$ and observing $y$, these beliefs are updated through the posterior $p(\theta\mid y,x)$. The informativeness of a potential design $x$ is then quantified by its expected information gain (EIG):

\begin{align}
\mathrm{EIG}(x)
&=
\mathbb{E}_{p(\theta)p(y \mid \theta,x)}
\left[
\log p(\theta \mid y,x)
-
\log p(\theta)
\right]
\label{eq:eig_posterior}
\\
&=
\mathbb{E}_{p(\theta)p(y \mid \theta,x)}
\left[
\log p(y \mid \theta,x)
-
\log p(y \mid x)
\right]
\label{eq:eig_likelihood}
,
\end{align}
which measures how much information design $x$ will provide about $\theta$, in expectation over possible $\theta$ and observations $y$. Using Bayes' rule, the EIG can be seen as either the expected reduction in uncertainty about $\theta$ after observing $y$ (Eq.~\ref{eq:eig_posterior}) or the expected reduction in uncertainty about $y$ from knowing $\theta$ (Eq.~\ref{eq:eig_likelihood}).

The optimal one-step design is defined as $x^\star = \arg\max_{x \in \mathcal{X}} \mathrm{EIG}(x)$, where $\mathcal{X}$ denotes the space of designs. This formulation extends naturally to the sequential setting where the design at step $t$ is chosen based on the history of previous experiments, $h_{t-1} = (x_1, y_1, \ldots, x_{t-1}, y_{t-1})$ by maximizing the EIG under the updated belief $p(\theta \mid h_{t-1})$. The experimental design at step $t$ can then be selected as $x_t^\star = \arg\max_{x \in \mathcal{X}} \mathrm{EIG}(x;h_{t-1})$.

\textbf{Challenges of Design Selection.}
Selecting designs according to the EIG in adaptive settings presents two major challenges. 
First, optimizing the EIG requires a costly estimation of the posterior $p(\theta \mid x_{1:t}, y_{1:t})$ at each step, which depends on the observed interaction history and therefore cannot be computed in advance. 
Moreover, even given this posterior, calculating the EIG is doubly intractable: it involves an outer expectation over possible latent states and observations, and an inner marginalization over latent states to compute the posterior predictive distribution $p(y_t \mid x_t, h_{t-1})$. 
Repeating this estimation and optimization procedure after every observation can be prohibitively expensive for settings that require decisions to be made in real time. 
Second, one-step EIG optimization considers only the immediate next design. This can be suboptimal in non-myopic decision states, where acquiring more information in the future requires first selecting a design with limited immediate information gain.

\subsection{Amortizing Sequential Experimental Design}
\label{sec:background_amort}
To address the computational cost of optimizing the EIG at inference time, Deep Adaptive Design (DAD, ~\citet{foster2021deep}) amortizes sequential experimental design into a design policy $\pi_\phi(x_t \mid h_{t-1})$, which maps the history $h_{t-1}$ to the next design $x_t$.
This shifts the computational cost to training, allowing the learned policy to directly select the next design at inference time without EIG optimization. 

Under this design-policy perspective, the standard sequential BED procedure described above corresponds to an implicit policy $\pi_S$, which selects each design by updating the posterior and optimizing the one-step EIG~\citep{foster2021deep}. Amortizing this policy directly would therefore require repeatedly simulating this expensive sequential BED procedure to generate training targets. Instead, DAD observes that the total EIG of a sequential experiment is equal to the sum of its conditional EIGs across steps. Summing these conditional information gains 
gives the information gained between the initial prior and the posterior after the complete trajectory. Applying Bayes' rule then expresses this quantity directly in terms of trajectory likelihoods:
\[
I_T(\pi)
=
\mathbb{E}_{p(\theta)p(h_T\mid\theta,\pi)}
\left[
\log p(h_T\mid\theta,\pi)
-
\log p(h_T\mid\pi)
\right],
\]
where $p(h_T\mid\pi)
=
\mathbb{E}_{p(\theta)}
\left[
p(h_T\mid\theta,\pi)
\right].$ 
DAD introduces the sequential prior contrastive estimation (sPCE) objective, a lower bound on the total EIG:
\begin{equation}
\label{eq:total_eig_spce}
\mathrm{EIG}_T(\pi)
=
\mathbb{E}_{\theta,h_T}\!\left[
\log\frac{p(h_T\mid\theta,\pi)}{p(h_T\mid\pi)}
\right]
\geq
\mathrm{sPCE}_T(\pi)
=
\mathbb{E}_{\theta_{0:L},h_T}\!\left[
\log
\frac{(L+1)\,p(h_T\mid\theta_0,\pi)}
{\sum_{\ell=0}^{L}p(h_T\mid\theta_\ell,\pi)}
\right],
\end{equation}
where $\theta_{0:L}\sim p(\theta)$ form a contrastive set used to approximate the marginal trajectory likelihood. The bound becomes tight as $L\to\infty$, making this design choice a trade-off between estimator bias and computational cost. Importantly, these hypotheses are sampled from the prior rather than from a history-dependent posterior, and therefore do not need to be re-estimated at each experimental step. 

\citet{blau2022optimizing} formulate the sequential design problem as an MDP by decomposing the sparse trajectory-level sPCE objective into per-step rewards:
\begin{equation}
\label{eq:spce_reward}
r_t
=
\log p(y_t \mid \theta_0, x_t)
-
\log \sum_{\ell=0}^{L} C_{t,\ell}
+
\log \sum_{\ell=0}^{L} C_{t-1,\ell},
\end{equation}
where $C_{t,\ell} = \prod_{k=1}^{t} p(y_k\mid\theta_\ell,x_k)$ is the likelihood of the observed outcomes up to step $t$ under hypothesis $\theta_\ell$, assuming observations are conditionally independent of the previous history given the hypothesis and current design. They then use a learned critic to estimate the expected future sPCE return, enabling non-myopic policy optimization.

\section{Related Work: BED within Large Language Models}
\label{sec}
\input{tables/related_works}
BED-based information gathering with LLMs has been studied through inference-time design selection and training-time policy optimization. Inference-time methods estimate EIG using LLM-induced generative models and select among candidate questions~\citep{kobalczyk2025active,choudhury2025bedllm}; CA-BED extends this approach to non-myopic selection through conversation-tree planning~\citep{arnould2026cabed}. ASIG~\citep{hartmann2026amortising} instead amortizes design selection into an LLM policy, while retaining a generative model derived from the policy LLM's predictive distributions.

Our approach explicitly optimizes the reasoning preceding design selection, constructs the training generative model independently of the policy LLM, and uses expected information rewards over multiple steps to allow for non-myopic information gathering. Table~\ref{tab:bed-comparison} summarizes the modeling and training choices across prior methods and our ablations. We discuss related works in detail in Appendix~\ref{app:related_work}.

\section{Method: Amortizing BED in Reasoning LLMs}
\label{sec:method}

Our goal is to amortize sequential BED into a reasoning LLM. One approach to using an LLM as a policy is to directly sample a design given an interaction history, $\pi_\phi(x_t\mid h_{t-1})$, and optimize its expected information gain~\citep{hartmann2026amortising}. Instead, our aim is to train an LLM that can \textit{reason through limited information}: deliberating over which hypotheses remain plausible, how possible observations distinguish them, and which further experiments those observations may enable.
We therefore parameterize the policy as
\begin{equation}
\pi_\phi(z_t,x_t \mid h_{t-1})
=
\pi_\phi(z_t \mid h_{t-1})
\cdot
\pi_\phi(x_t \mid h_{t-1},z_t).
\label{eq:reasoning_design_policy}
\end{equation}
and train its joint output $(z_t,x_t)$ with an information gain objective. 
At a high level, our training procedure consists of three steps:
\begin{enumerate}
\item \textbf{Generate a design through reasoning.} Given an interaction history $h_{t-1}$, the policy generates a reasoning trace $z_t$ followed by a design $x_t$.
\item \textbf{Evaluate its information value.} A generative model, specified independently of the policy, provides the hypothesis and outcome distributions used to compute the information gain of the design $x_t$.
\item \textbf{Optimize the reasoning policy.} We use information-gain rewards accumulated over $K$-step rollouts to train the reasoning policy with Group Relative Policy Optimization (GRPO)~\citep{shao2024deepseekmath}.
\end{enumerate}

This procedure requires two main design choices: how to construct the generative model used to evaluate information gain (Sec.~\ref{sec:DQ2}), and how to turn this information signal into a training reward (Sec.~\ref{sec:DQ3}).

\subsection{How should the generative model $p(\theta,y ; x)$ be constructed?}
\label{sec:DQ2}
\label{sec:generative-model}
Estimating the EIG requires specifying a generative model $p(\theta,y;x)$ of the experiment. In our amortized setting, this choice determines both the quality of the training signal and the computational cost of obtaining it. We build on the contrastive formulation in Sec.~\ref{sec:background_amort}, which requires specifying two components: a prior $p(\theta)$ over contrastive hypotheses and a likelihood $p(y\mid\theta,x)$ for evaluating proposed designs. We describe how we construct these components below.

\textbf{Decoupling the generative model from the policy.}
One option is to use the policy LLM itself to estimate both components, as in the approaches discussed in Sec.~\ref{sec}. Training-time amortization enables an important flexibility: the generative model can instead be specified independently of the policy being trained, without requiring it for design selection at inference time. This allows us to define the experimental distribution used for training upfront, while keeping it fixed as the policy is updated (see Appendix~\ref{app:external_generative_model}). This is particularly useful when the initial policy provides a poor approximation of that distribution. For example, the training prior can place mass on hypotheses that the policy itself would be unlikely to generate, while a separate likelihood model can provide more reliable estimates of $p(y\mid\theta,x)$ than the policy's own predictive distribution.

\textbf{The prior $p(\theta)$.}
The contrastive hypotheses can be sampled from the prior for each trajectory, as in sPCE, or fixed across training. We define our prior as a uniform distribution over a finite contrastive set $\Theta_C=\{\theta_\ell\}_{\ell=0}^{L}$, with $p_C(\theta_\ell)=1/(L+1)$. This prior is fixed across all training trajectories and defines the hypothesis distribution with respect to which information gain is computed. The size of the set introduces a trade-off between coverage and computational cost, while its composition and prior weights determine which hypotheses the policy is trained to distinguish. 

\textbf{The likelihood $p(y\mid\theta,x)$.}
In an open-ended natural-language design space, the possible designs $x$ cannot generally be enumerated in advance, making it impractical to precompute their likelihoods. We therefore use a separate, larger LLM as a semantic likelihood evaluator $p_{\mathrm{oracle}}(y\mid\theta,x)$, which can be queried during training. Following~\citet{blau2022optimizing}, we assume that each observation is conditionally independent of the interaction history given the hypothesis and current design, allowing us to accumulate likelihoods as $C_{t,\ell}$ (see Sec.~\ref{sec:background_amort}). The observation $y$ may be stochastic even when $\theta$ is known. For example, the answer to ``Is it cute?'' is ambiguous for a fixed $\theta$~\citep{choudhury2025bedllm}. We therefore retain the full likelihood distribution over the finite outcome space $\mathcal{Y}$. Each possible outcome $y$ induces a corresponding updated likelihood, $C_{t,\ell}^{(y)} = C_{t-1,\ell}p_{\mathrm{oracle}}(y\mid\theta_\ell,x_t)$, which allows us to marginalize over possible outcomes when computing the reward (discussed in Sec.~\ref{sec:DQ3}). This design also allows for likelihood evaluations $p_{\mathrm{oracle}}(y\mid\theta_\ell,x_t)$ to be cached and reused across training.

\subsection{How should the reward be assigned?}
\label{sec:DQ3}
\label{sec:information-returns}

Having specified the generative model in Sec.~\ref{sec:DQ2}, we now consider how information gain over the training hypothesis space should be converted into a reward and how these rewards should be aggregated into training returns. 

\textbf{Computing the Reward.}
A natural choice is to use the realized incremental sPCE reward in Eq.~\ref{eq:spce_reward}~\citep{blau2022optimizing}, which can be written as
\begin{equation}
r_t^{\mathrm{real}}
=
\log\frac{p_{\mathrm{oracle}}(y_t\mid\theta^\star,x_t)}
{p_C(y_t\mid h_{t-1},x_t)},
\label{eq:realized_information_reward}
\end{equation}
where $p_C(y\mid h_{t-1},x_t)=\sum_{\ell=0}^{L}p_C(\theta_\ell\mid h_{t-1})p_{\mathrm{oracle}}(y\mid\theta_\ell,x_t)$ denotes the posterior predictive distribution and $p_C(\theta_\ell\mid h_{t-1})=\frac{C_{t-1,\ell}}{\sum_j C_{t-1,j}}$ is the posterior over the fixed contrastive hypothesis space under the uniform prior (see Appendix~\ref{app:fixed_contrastive} for the full derivation).

For a sampled target $\theta^\star$ and observed outcome $y_t$, this assigns the proposed design its realized information gain. When $\theta^\star\sim p_C(\cdot\mid h_{t-1})$ and $y_t\sim p_{\mathrm{oracle}}(\cdot\mid\theta^\star,x_t)$, the expectation of this reward is exactly the one-step EIG under the current finite belief.

Because our hypothesis and outcome spaces are finite and enumerable, we can bring this expectation inside each training step: rather than relying on a single realized $(\theta^\star,y_t)$ pair, we marginalize over both hypotheses and outcomes at each turn. We then assign the expected information reward
\begin{equation}
r_t^{\mathrm{EIG}}
=
\sum_{\ell=0}^{L} p_C(\theta_\ell\mid h_{t-1})
\sum_{y\in\mathcal{Y}}
p_{\mathrm{oracle}}(y\mid\theta_\ell,x_t)
\left[
\log p_{\mathrm{oracle}}(y\mid\theta_\ell,x_t)
-
\log p_C(y\mid h_{t-1},x_t)
\right].
\label{eq:expected_eig_reward}
\end{equation}
Under the sampling assumptions above, $r_t^{\mathrm{EIG}}=\mathbb{E}[r_t^{\mathrm{real}}\mid h_{t-1},x_t]$. This is the Rao--Blackwellized form of the realized reward, removing the variance associated with sampling a particular $(\theta^\star,y_t)$ pair at a given history and design. The resulting reward is exactly the one-step EIG over the specified finite hypothesis space, though not necessarily over the full hypothesis space of interest (see Appendix~\ref{app:realized_expected}).

In practice, we may want to use a large contrastive set to provide a rich information-gain signal, but generate training trajectories from only a smaller subset of hypotheses for computational efficiency. In this case, realized and expected rewards can differ beyond variance reduction, as we discuss further in Appendix~\ref{app:target_mismatch}.

\textbf{Constructing the Return.}
The reward in Eq.~\ref{eq:expected_eig_reward} measures only the immediate information value of a design. However, a design may also enable more informative future interactions. Following the sequential formulation of \citet{blau2022optimizing}, the undiscounted sum of per-step information rewards recovers multi-step EIG in expectation when rollouts follow the specified generative model. We therefore evaluate each proposed design through a rollout of the current policy. We limit the horizon to $K$ steps for tractability and introduce $\gamma$ to discount future information:
\begin{equation}
R_{t,K,\gamma}
=
\sum_{k=0}^{K-1}\gamma^k r_{t+k}^{\mathrm{EIG}}.
\label{eq:k_step_return}
\end{equation}
Setting $K=1$ recovers the myopic objective, while $K>1$ with $\gamma>0$ rewards the initial decision for information acquired at subsequent steps. 
In practice, our TQ implementation prunes low-weight hypotheses and advances rollouts using the oracle’s most likely answer rather than sampling an outcome. We discuss these approximations to the fully stochastic objective further in Appendix~\ref{app:information-returns}.

\section{Experiments}
\label{sec:experiments}
We now evaluate our approach for amortizing sequential BED in a reasoning LLM in four studies:

\begingroup
\arrayrulecolor{myBlue}
\setlength{\arrayrulewidth}{1.2pt}

\begin{table}[h]
\centering
\footnotesize

\begin{tabular}{|p{0.06\linewidth}|p{0.80\linewidth}|}
\hline
\rowcolor{myBlue}
\textcolor{white}{\textbf{Section}} &
\textcolor{white}{\textbf{Question}} \\
\hline

\S\ref{sec:exp_reason} &
Does amortizing BED into a reasoning LLM improve general information acquisition? \\
\hline

\S\ref{sec:exp_gen} &
How does our approach compare to existing inference- and training-time BED approaches? \\
\hline

\S\ref{exp:ablations} &
How do choices for estimating and optimizing information gain affect performance? \\
\hline

\S\ref{sec:task_generalization} &
Does a policy trained for information acquisition in one task transfer to a different task? \\
\hline

\end{tabular}

\label{tab:experiment-roadmap}
\end{table}

\arrayrulecolor{black}
\endgroup

\textbf{Training Environment}
Following prior work~\citep{kobalczyk2025active,choudhury2025bedllm,hartmann2026amortising}, we use Twenty Questions (TQ), where an agent must identify a hidden target within 20 questions. Here, $\theta$ is the target (e.g., ``horse''), $x$ is an open-ended natural-language question (e.g., ``Is it a mammal?''), and the observation lies in the finite answer space $\mathcal{Y}=\{\text{yes},\text{no},\text{maybe}\}$. We use \texttt{GPT-5} as the semantic likelihood evaluator $p(y|\theta,x)$.

\textbf{Training Data}
Our main experiments train on the \texttt{Animals} domain. We use a fixed contrastive set of 200 animals for computing the training reward, of which 100 are used as ground-truth targets $\theta$ when constructing training trajectories. We use \texttt{GPT-5} to generate 250 Twenty Questions games over these 100 targets. From each game, we sample up to five intermediate points before the final guess, yielding 1,200 partial trajectories. Each sampled trajectory defines a history $h_{t-1}$ containing the sequence of questions and answers observed up to that point. The task-transfer experiment uses a separate training dataset, described in Sec.~\ref{sec:task_generalization}.

\textbf{Evaluation Data.}
Each model plays 50 Twenty Questions games across three seeds, with a different target sampled for each game in the corresponding evaluation domain. We evaluate on held-out animals drawn from the same 200-animal contrastive set used during training. 
In Sec.~\ref{exp:ablations}, we consider animals that are absent from both the training targets and the training contrastive set. To assess cross-domain transfer, we evaluate on \texttt{Plants}, \texttt{Celebrities}, and \texttt{Countries}. Sec.~\ref{sec:task_generalization} evaluates task transfer by applying models trained on Twenty Questions to \textsc{mediQ}~\citep{li2024mediq}, an interactive medical question-asking benchmark.

\textbf{Models and Hyperparameters}
We call our method \emph{Amortized Reasoning for Bayesian Experimental Design} (AR-BED). Unless otherwise specified, $\mathrm{AR\text{-}BED}_{K}$ is a \texttt{Gemma-4-E4B-it} reasoning policy trained with expected information rewards with rollout horizon $K$. We denote by $\mathrm{AR\text{-}BED}_{K}^{\mathrm{Direct}}$ the ablation where we train without reasoning, directly sampling and rewarding a question from the history $x\sim p(.|h)$. We use a batch size of 128 with 8 rollouts per state, and train for 250 steps. For multi-step training ($K>1$), we set $\gamma=0.5$. Full training details are in Appendix~\ref{app:experiment-details}.

\textbf{Metrics.}
We report the solve rate within a 20-question budget and the average EIG per turn to account for varying game lengths.

\subsection{Amortizing BED into Reasoning Improves General Information Acquisition}
\label{sec:exp_reason}
\label{sec:h1}
\begin{figure}[!t]
    \centering
    \includegraphics[width=\linewidth]{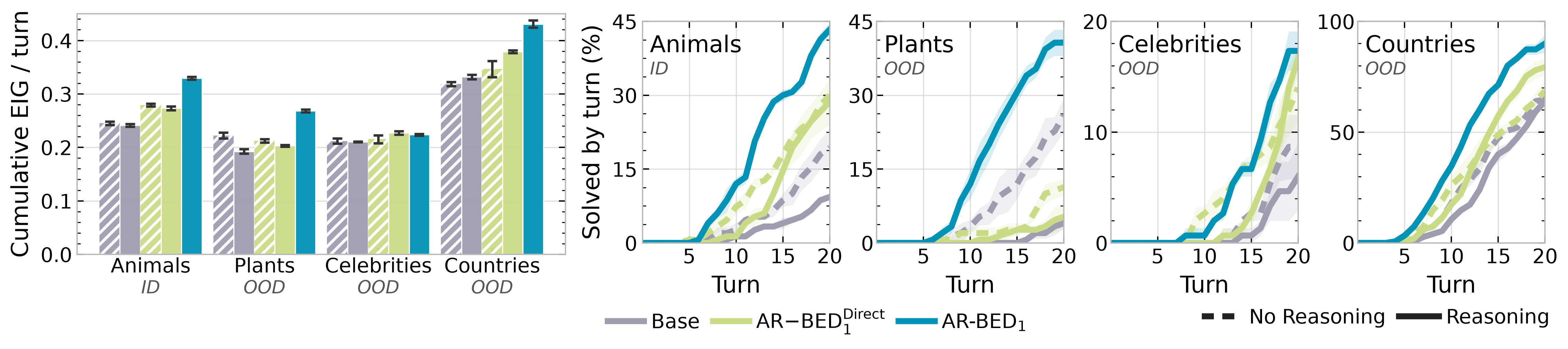}
    \caption{\textbf{Amortizing BED into reasoning improves information acquisition and transfer.} 
    Bars and bands show mean $\pm$ SEM over three evaluation repeats.
    }
    \label{fig:reasoning}
\end{figure}

\textbf{Setup.}
We evaluate whether amortizing BED into the reasoning process improves information acquisition both in- and out-of-distribution. We compare our model, $\mathrm{AR\text{-}BED}_{1}$, to four controls: the \textsc{Base} model and the directly trained $\mathrm{AR\text{-}BED}_{1}^{\mathrm{Direct}}$ model, each evaluated with and without reasoning at inference time. This design separates gains from BED training from gains due only to enabling reasoning at test time.

\textbf{Results.} (Fig.~\ref{fig:reasoning})
Adding reasoning to the \textsc{Base} model \textit{reduces} final solve rate across all categories and increases average EIG only on Countries. This suggests that inference-time reasoning alone does not improve information gathering, and may even worsen it.\footnote{A similar effect is observed for some models reported in Table~1 of~\citet{choudhury2025bedllm}.} Training a non-reasoning policy improves the in-distribution behavior but the transfer to out-of-distribution domains is inconsistent: it is worse than \textsc{Base} on Plants but improves on Celebrities and Countries. This suggests that direct amortization can learn useful in-distribution question-sampling behavior, but does not
produce robust transfer across domains. In contrast, our model $\mathrm{AR\text{-}BED}_{1}$ achieves the best final solve rate across all categories and obtains the highest average EIG per turn on Animals, Plants, and Countries. These gains over both the \textsc{Base} model and the direct policy suggest that training a reasoning model may enable general information-gathering strategies.

\subsection{Amortizing BED into reasoning outperforms Inference- and Training-Time BED Baselines} \label{sec:exp_gen}

\textbf{Setup.}
We next compare against BED-style alternatives that differ in how they obtain the generative model used to estimate EIG. Our $\mathrm{AR\text{-}BED}_{1}$ model uses an external generative model, whereas \textsc{BED-LLM}~\citep{choudhury2025bedllm} uses the LLM itself to construct the predictive generative model at inference time. We also compare to \textsc{ASIG}, a training-time amortization baseline, which uses the same BED-LLM-style LLM-generated model during training but trains a direct question-sampling policy rather than a reasoning policy~\citep{hartmann2026amortising}. To isolate the effect of the information-gain objective, we do not include an additional game reward for \textsc{ASIG}. We evaluate \textsc{ASIG} with and without reasoning at inference.

\begin{wraptable}{r}{0.38\linewidth}
\vspace{-1.0em}
\centering
\small
\setlength{\tabcolsep}{3pt}
\renewcommand{\arraystretch}{0.95}
\begin{tabular}{@{}p{0.34\linewidth}lr@{}}
\toprule
Setting & Model & Time \\
\midrule
Evaluation & BED-LLM & 2376 s \textbf{(61.1$\times$)} \\
(game) & AR-BED$_1$ & 38.9 s \\
\midrule
Training & ASIG & 136.3 s \textbf{(2.9$\times$)} \\
(step) & AR-BED$_1$ & 47.3 s \\
\bottomrule
\end{tabular}
\caption{\textbf{$\mathrm{AR\text{-}BED}_{1}$ is more efficient at both inference and training time.}}
\label{tab:tq-compute-time}
\vspace{-1.0em}
\end{wraptable}

\begin{figure}[t]
    \centering
    \includegraphics[width=\linewidth]{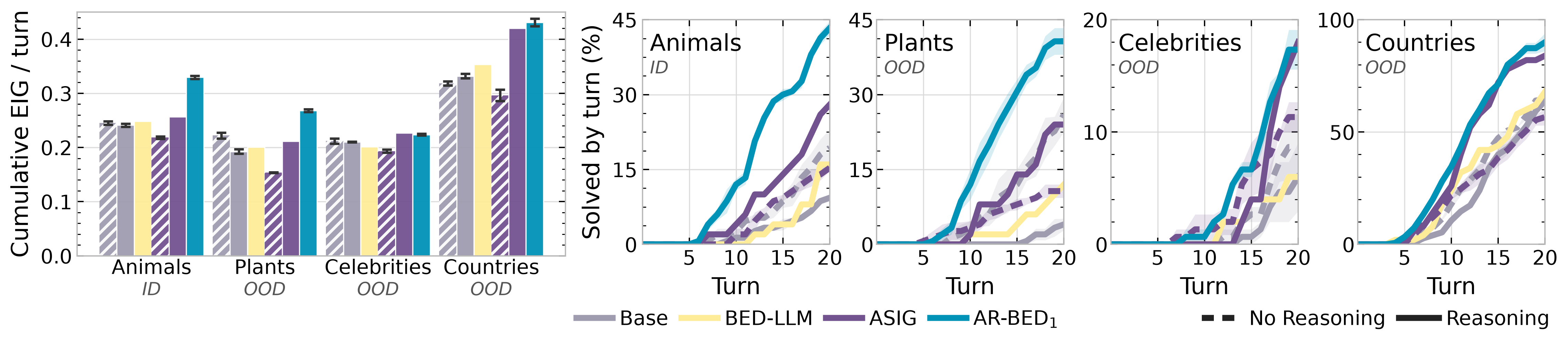}
    \caption{\textbf{Comparison to BED baselines.} $\mathrm{AR\text{-}BED}_{1}$ outperforms inference-time \textsc{BED-LLM} and training-time \textsc{ASIG} on most domains; bars and bands show mean $\pm$ SEM over evaluation repeats.}
    \label{fig:generative}
\end{figure}

\textbf{Results.}
Figure~\ref{fig:generative} shows that $\mathrm{AR\text{-}BED}_{1}$ outperforms the inference-time and training-time BED baselines on most domains. \textsc{BED-LLM} improves only modestly over the reasoning \textsc{Base} model, and is worse than the non-reasoning \textsc{Base} model on some domains in both solve rate and EIG. This suggests that online BED with an LLM-generated model of the experiment may be limited by the model's ability to represent the relevant distributions $p(\theta)$ and $p(y \mid \theta, x)$. Similarly, \textsc{ASIG} without reasoning does not improve over the non-reasoning \textsc{Base} model, indicating that directly amortizing this LLM-generated BED objective suffers from similar limitations. Adding reasoning at inference substantially improves \textsc{ASIG}, but it still remains below $\mathrm{AR\text{-}BED}_{1}$ on Animals, Plants, and Countries, and is comparable on Celebrities. These results suggest that the benefit comes from amortizing the information-gain objective into the model's reasoning process and using an external generative model that provides a more reliable training signal.

Table~\ref{tab:tq-compute-time} shows that at inference time, \textsc{BED-LLM} is about $61\times$ slower per game than $\mathrm{AR\text{-}BED}_{1}$ because it performs BED computation online. At training time, \textsc{ASIG} is about $2.9\times$ slower per step due to the additional cost of using the LLM-generated BED-LLM-style generative model. Thus, $\mathrm{AR\text{-}BED}_{1}$ gains stronger information acquisition capabilities while avoiding the main computational costs of BED methods with a policy-based generative model.

\begin{wrapfigure}{r}{0.40\linewidth}
    \vspace{-1.5em}
    \centering
    \includegraphics[width=\linewidth]{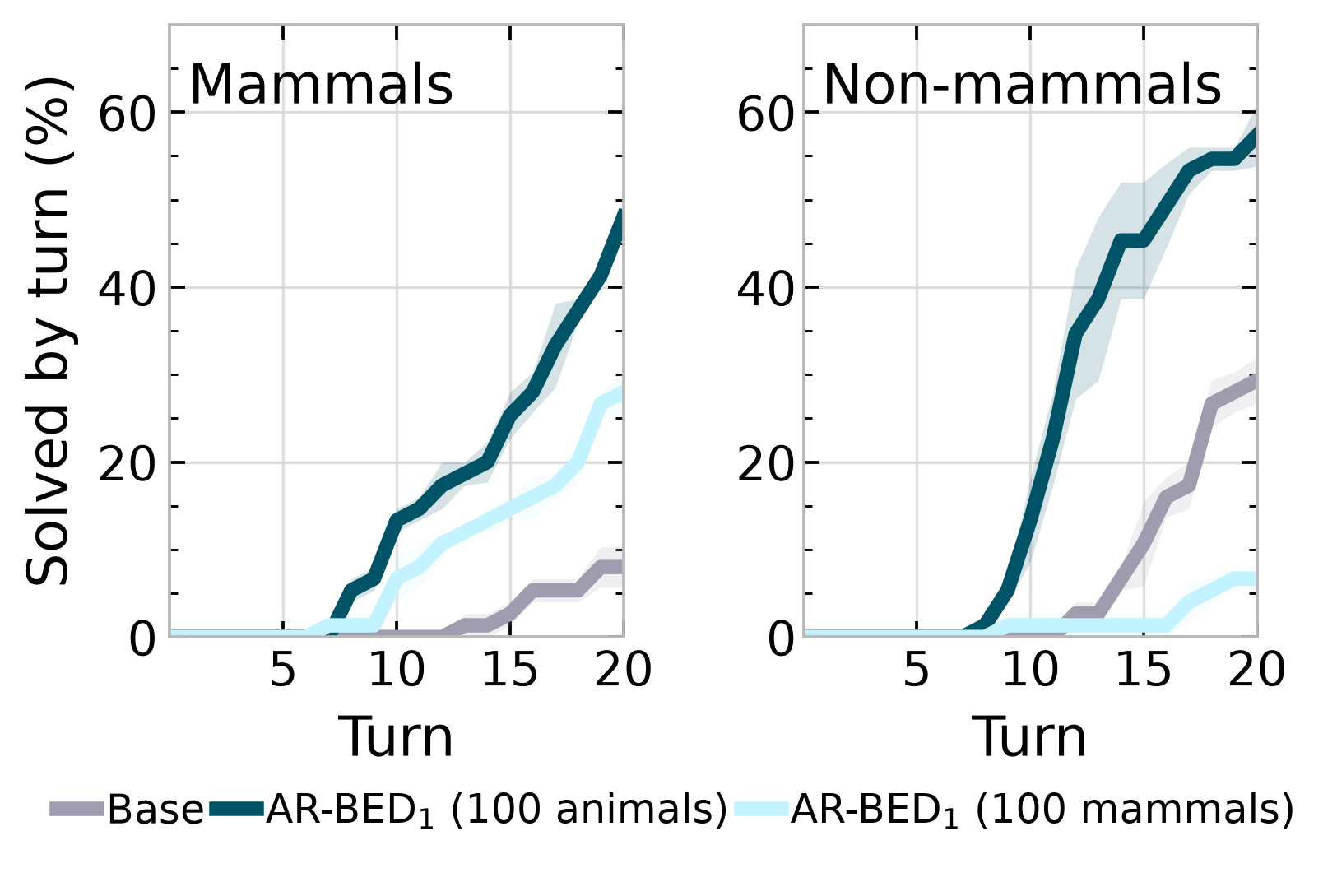}
    \caption{\textbf{Generalization across held-out animal subgroups.}
    Solve rate over turns for held-out mammals and non-mammals. Bands show mean $\pm$ SEM over three evaluation repeats.}
    \label{fig:mammals}
    \vspace{-1em}
\end{wrapfigure}
\subsection{Ablations and Insights}
\label{exp:ablations}

\subsubsection{Misspecified Priors and Size of the Contrastive Set}
\label{exp:misspecified_priors}
To study the effect of a misspecified prior (i.e., training a model with a flawed contrastive set), we compare two variants of $\mathrm{AR\text{-}BED}_{1}$: one with a 100-animal contrastive set trained on only these animals, and one with a 100-mammal contrastive set trained only on these mammals. Both models are prompted to guess in the category \texttt{Animals}. As a result, the mammals-trained model received a flawed information signal compared to the true hypothesis space $p(\theta)$: for example, the question ``Is it a mammal?'' receives $EIG\approx0$ despite being a strong question to discriminate between animals.

Fig.~\ref{fig:mammals} shows the solve rate of both models on held-out sets of 25 mammals and 25 non-mammals. The model trained with a mammals-only contrastive set performs significantly worse than both the animals-trained and base models on non-mammals, suggesting that it learned a flawed representation of the animals hypothesis space. Notably, it also leads to worse performance than the animals-trained model on mammal targets. This suggests that training on the broader hypothesis space of animals might provide exposure to a diversity of features that is beneficial even within the shared mammal subset.

We additionally investigate the effect of the size of the contrastive set in Appendix~\ref{app:extended_results:contrastive_ablations}, finding that using a 200-sample contrastive set improves the OOD transfer performance compared to a 100-sample contrastive set.

\subsubsection{Using expected or realized IG in the reward signal}
\label{exp:reward_estimator}
Marginalizing over hypotheses and answers generally provides a stronger training signal than using the realized information gain: expected-reward training improves both EIG and solve rate on Animals, Plants, and Countries, although not on Celebrities. Appendix~\ref{app:extended_results:reward_estimator} reports the matched comparison, consistent with the variance-reduction motivation in Sec.~\ref{sec:information-returns}.

\subsubsection{Using a non-myopic reward signal}
\label{exp:nonmyopic}
Appendix~\ref{app:extended_results:nonmyopic} brings together three complementary evaluations of multi-step rewards. In a natural-language adaptation of Cube-NM~\citep{schütz2026afabenchgenericframeworkbenchmarking}, the two-step policy learns to acquire initially uninformative context for its downstream benefit, substantially outperforming the one-step policy in later disease identification. On the TwentyQuestions target-prediction task of \citet{wang2025adaptive}, multi-step policies improve attribute prediction, particularly on medium and hard questions. Finally, we empirically investigate the extent to which Twenty Questions as played by the GPT-5 policy shows signs of non-myopicity. We show that immediate- and downstream-EIG rankings can differ, while matched-state evaluations find some gains from $K=2$ and $K=3$ over $K=1$ at two- and three-question horizons. Together, these results distinguish a controlled non-myopic mechanism from the smaller benefits of lookahead in open-ended questioning.

\subsection{Generalization to other tasks}
\label{sec:task_generalization}
The preceding experiments examine transfer across domains within TQ. We now
ask whether the learned information-gathering behavior also transfers
\textit{to a different task}. We use MediQ~\citep{li2024mediq}, which simulates
clinical questioning: given partial patient information and a multiple-choice
medical inquiry, a policy asks open-ended questions to gather evidence from
an LLM-based Patient with full access to patient records.

\textbf{Setup.} We construct three synthetic medical TQ environments---
dermatology, neurology, and gastroenterology---each with a 50-disease
contrastive set drawn from MediQ's answer inventory. Using the
$\mathrm{AR\text{-}BED}_{1}$ objective, we train a separate medical policy on
1,200 starting states and evaluate its checkpoint against the
base model. Both policies ask five questions with GPT-5 as the fixed Patient.
We evaluate the quality of each model's line of questioning by using a
separate GPT-5 decider to predict the answer from the initial information and interaction history. 

We report two diagnosis subsets: 95 dermatology cases whose correct diagnoses
are absent from the training contrastive set (\textit{same domain, different
task, held-out labels}), and 92 cases from other specialties, including
pediatrics, psychiatry, and cardiology (\textit{different domain and task}).
Full training and evaluation details are in Appendix~\ref{app:mediq}.

\input{tables/mediq}

\textbf{Results.} Table~\ref{tab:mediq-transfer} shows that medical
$\mathrm{AR\text{-}BED}_{1}$ improves accuracy from $90.5\%$ to $94.7\%$ on
held-out dermatology diagnoses and from $84.8\%$ to $90.2\%$ on other
specialties. It also increases the probability assigned to the correct answer
and reduces the Brier score in both domains. These results indicate that the
reasoning capabilities learned by $\mathrm{AR\text{-}BED}_{1}$ in a synthetic
TQ environment transfer to a realistic, simulated clinical dialogue setting,
including medical specialties not used during fine-tuning.

\section{Conclusion}

We studied how sequential Bayesian experimental design can be amortized into a reasoning LLM to improve adaptive information gathering. Our approach makes two key design choices: decoupling the experimental generative model from the policy to avoid relying on potentially poor predictive distributions, and constructing expected information-gain rewards that can be accumulated across interactions. We find that our method improves information acquisition across unseen hypothesis spaces, outperforming other methods, and that the learned behavior transfers to Medi-Q, a real-world clinical dialogue dataset. Our results suggest that adaptive information gathering can be learned as a transferable reasoning capability in LLMs rather than reconstructed through explicit search at every interaction.

\textbf{Limitations.} We study relatively short rollout horizons. Extending the approach to longer horizons while maintaining a reliable training signal is an important direction. We also find that performance is sensitive to the coverage of the contrastive hypothesis set when it is misspecified relative to the true hypothesis space. Further understanding how the size, composition, and prior weighting of the contrastive set affect learning and transfer is therefore an important avenue for future study.

% \clearpage
\section*{AI Use Statement}

In this work, we used generative AI tools to assist with the coding implementation, figure editing, LaTeX formatting, and manuscript polishing. The authors determined the research questions, methodology, experiments, and final interpretation of results. We reviewed all AI-assisted work and take responsibility for the final content of this paper.

\section*{Acknowledgements}

AK’s PhD studentship is funded by Sanofi, while AB’s studentship is funded by Eedi.
This work was supported by Azure sponsorship credits granted by Microsoft’s AI for Good Research Lab.

\clearpage
\bibliography{references}
\bibliographystyle{iclr2027_conference}

\newpage
\appendix

\section{Extended Related Works}
\label{app:related_work}
\input{sections/related_works.tex}

\section{Extended Experimental Results}
\label{app:extended_results}

\subsection{Contrastive set ablations}
\label{app:extended_results:contrastive_ablations}
\begin{figure}[h!]
    \centering
    \includegraphics[width=\linewidth]{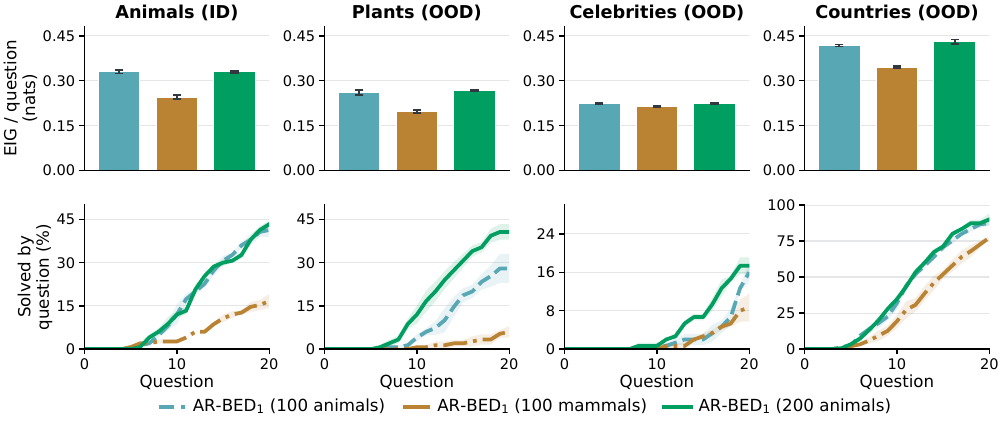}
    \caption{\textbf{Training-distribution size and coverage affect acquisition and transfer.} 
    }
    \label{fig:contrastive-coverage}
\end{figure}

\textbf{A larger contrastive set improves performance.}
Figure~\ref{fig:contrastive-coverage} compares 100- and 200-animal training generative models. On Animals, their EIG per question is nearly identical ($0.330$ nats), while the 200-animal policy has a slightly higher final solve rate ($43.3\%$ versus $41.3\%$). The clearest difference is on Plants, where the larger set improves EIG from $0.260$ to $0.268$ and solve rate from $28.0\%$ to $40.7\%$. It also improves country EIG and solve rate; celebrity EIG is essentially unchanged. These results suggest that increasing the size of the contrastive set can improve the usefulness of the training signal and its transfer, but does not guarantee better acquisition on every domain.

\textbf{Coverage matters beyond the number of hypotheses.}
Restricting the training generative model to 100 Mammals substantially reduces performance relative to 100 diverse Animals. On animal evaluation targets, EIG falls from $0.330$ to $0.245$ and solve rate from $41.3\%$ to $16.7\%$; both metrics are also lower on all three OOD domains. Since this condition restricts both the contrastive set and sampled training targets, it tests generative-model coverage rather than a change to the estimator alone. The result reinforces the design choice in Sec.~\ref{sec:generative-model}: decoupling the generative model from the policy allows its coverage to be controlled, but the resulting information reward is only as useful as the distinctions represented during training.

\subsection{Expected versus realized information rewards}
\label{app:extended_results:reward_estimator}
\textbf{Setup.}
We compare one-step reasoning policies trained on the same sampled animal-game states with the same 200-animal contrastive set. The policies differ only in whether the reward marginalizes over hypotheses and possible answers or uses the information gain realized along the sampled trajectory. Both policies generate a reasoning trace before selecting a question; this comparison changes the reward estimator, not which part of the output is optimized.

\begin{figure}[!htbp]
    \centering
    \includegraphics[width=\linewidth]{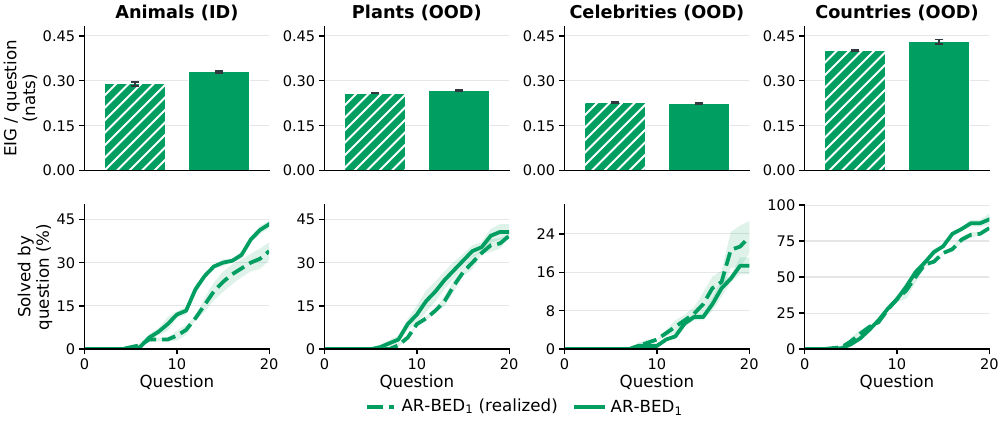}
    \caption{\textbf{Expected rewards generally improve learning over realized rewards.} One-step reasoning policies trained with the same 200-animal generative model, differing in reward estimator. Expected rewards improve EIG per question and solve rate on Animals, Plants, and Countries, but not Celebrities. Bars and bands show mean $\pm$ SEM over three evaluation repeats.}
    \label{fig:reward-estimator}
\end{figure}

\textbf{Expected information rewards usually provide the stronger training signal.}
Figure~\ref{fig:reward-estimator} compares information acquisition and solve rate on Animals and three transfer domains. Expected-reward training improves animal EIG from $0.290$ to $0.330$ nats per question and solve rate from $34.0\%$ to $43.3\%$, and improves both metrics on Plants and Countries. The exception is Celebrities, where realized-reward training yields slightly higher EIG and a higher solve rate ($23.3\%$ versus $17.3\%$). The overall pattern is consistent with the variance-reduction motivation in Sec.~\ref{sec:information-returns}, but does not establish variance as the sole cause or imply a uniform advantage across transfer domains.

\FloatBarrier
\subsection{Non-myopic information acquisition}
\label{app:extended_results:nonmyopic}
We evaluate multi-step rewards in three complementary ways: a controlled environment where context must be acquired before it becomes useful, prediction of target attributes during open-ended questioning, and information acquisition in Twenty Questions. The Cube-NM experiment isolates the ability to learn an initially uninformative action; the other evaluations examine whether a longer reward horizon also benefits natural-language policies.

\subsubsection{Cube-NM: acquiring context before information}
\label{sec:h2:cube}
\textbf{A controlled non-myopic environment.}
We adapt Cube-NM~\citep{schütz2026afabenchgenericframeworkbenchmarking,ruckstiess2013minimizing} as a small mechanism check. Its 32 diseases and three equally likely, disease-independent contexts define 96 disease--context pairs. Each context activates eight tests; a separate context test reveals which block is active but provides zero immediate disease information. A useful policy must therefore trade immediate information for better subsequent acquisition. We use a natural-language test-selection interface without reasoning traces to isolate this routing decision from open-ended question generation and medical knowledge.

\textbf{Training and evaluation.}
The 25 equal-cost tests comprise three context-dependent blocks of eight tests and one context test. The context is independent of the disease, and outcomes are deterministic given both. The textual interface returns positive, negative, or inconclusive outcomes and the diseases invalidated by the observation; the context test instead lists the informative tests. Test identifiers are randomly permuted once, identically across the matched runs, and the outcome matrix and context-test identity are not initially disclosed. The policy selects one identifier without a reasoning trace. We train \texttt{Qwen3-0.6B} with GRPO/LoRA for 250 updates at learning rate $10^{-5}$, sampling 16 states and 16 rollouts per state from a rolling on-policy bank covering turns 0--4. Temperature is 2, the entropy coefficient is 0.02, and $\gamma=0.9$. Evaluation uses greedy actions over all 96 disease--context pairs and an exact Bayesian MAP decoder. Total EIG is $\log 32$ minus the mean disease-posterior entropy. These are matched single-training-seed results, not estimates of variability across training runs.

\begin{figure}[!htbp]
    \centering
    \includegraphics[width=\linewidth]{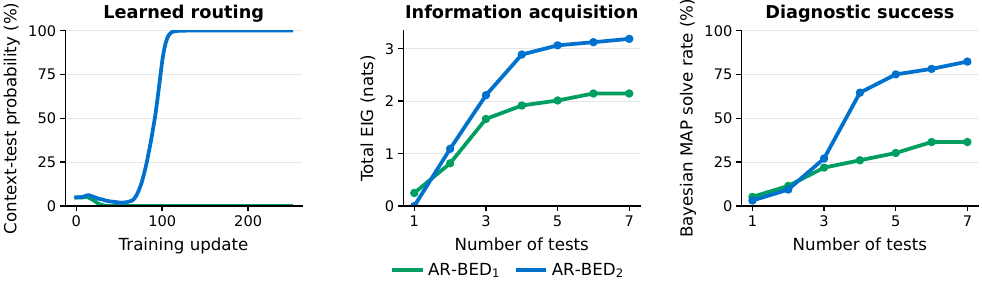}
    \caption{\textbf{A controlled confirmation of non-myopic learning in Cube-NM.} Left: probability of selecting the initially uninformative context test during training. Center: total EIG under greedy acquisition. Right: Bayesian MAP solve rate. The two-step objective learns to acquire context before disease information, improving later acquisition. These are matched single-seed, no-reasoning policies evaluated on all 96 disease--context pairs; the decoder is Bayesian, not an LLM diagnosis.}
    \label{fig:cube-nm-overview}
\end{figure}

\textbf{The two-step reward learns the initially uninformative context test.}
Figure~\ref{fig:cube-nm-overview} shows that the two-step policy learns to select the context test with almost unit probability, while the one-step policy favors an immediately informative disease test. This incurs an initial cost: after one test, Bayesian MAP solve rate is $3.1\%$ for $K=2$ versus $5.2\%$ for $K=1$. The two-step policy overtakes after three tests ($27.1\%$ versus $21.9\%$) and reaches $75.0\%$ versus $30.2\%$ after five. Thus the multi-step reward can learn a design for its downstream benefit, even when its immediate disease-information reward is zero. These matched single-seed results support the intended mechanism, not global optimality or an additional claim about reasoning transfer.

\subsubsection{Why Cube-NM requires non-myopic information acquisition}
\label{app:cube_nm_nonmyopic}

Cube-NM provides a controlled setting in which an action can have zero
immediate information gain about the target while increasing the value of a
subsequent experiment. We use this structure to illustrate why a one-step
information-gain objective cannot assign value to such routing actions, while
a two-step objective can.

\textbf{Setup.}
Let $\Theta$ denote the latent disease and let $C$ denote an auxiliary context
variable. In Cube-NM, the context is independent of the disease,
\[
\Theta \perp C.
\]
The disease tests are partitioned into context-dependent blocks
$\{\mathcal X_c\}_{c\in\mathcal C}$. Tests in $\mathcal X_c$ are informative
about $\Theta$ when the active context is $C=c$.

In addition, there is a context test $x^{\mathrm{ctx}}$ whose outcome reveals
the active context,
\[
Y^{\mathrm{ctx}} = C,
\]
but does not directly reveal the disease.

For a belief state $b$, define the immediate information value of a test $x$ as
\[
r(b,x)
=
I(\Theta;Y_x\mid b,x).
\]

\textbf{The context test has zero myopic information gain.}
Because $Y^{\mathrm{ctx}}=C$ and $\Theta\perp C$,
\[
r(b_0,x^{\mathrm{ctx}})
=
I(\Theta;C)
=
0,
\]
where $b_0$ denotes the initial belief.

By construction, there exist disease tests with positive immediate information
gain,
\[
\max_{x\neq x^{\mathrm{ctx}}}
I(\Theta;Y_x) > 0.
\]

Therefore the context test cannot be selected by a policy that strictly
maximizes one-step information gain:
\[
x^{\mathrm{ctx}}
\notin
\arg\max_x Q_1(b_0,x),
\qquad
Q_1(b,x)=I(\Theta;Y_x\mid b,x).
\]

Thus a myopic information-gain objective assigns no credit to the context test,
despite the fact that observing the context identifies which subsequent tests
are informative.

\textbf{A two-step objective assigns value to the context test.}
For a two-step policy, define
\[
Q_2(b,x)
=
I(\Theta;Y_x\mid b,x)
+
\gamma\,
\mathbb E_{y\sim p(y\mid b,x)}
\left[
V_1(b^{x,y})
\right],
\]
where
\[
V_1(b')
=
\max_{x'} I(\Theta;Y_{x'}\mid b',x')
\]
is the value of the best remaining one-step experiment, and $b^{x,y}$
denotes the belief state after observing outcome $y$ from test $x$.

For the context test, the first term is zero. However, observing its outcome
reveals $C$, and therefore identifies the relevant block of disease tests.
Consequently,
\[
Q_2(b_0,x^{\mathrm{ctx}})
=
\gamma
\mathbb E_{C}
\left[
\max_{x\in\mathcal X_C}
I(\Theta;Y_x\mid C)
\right].
\]

If every context contains at least one disease-informative test,
\[
\max_{x\in\mathcal X_c}
I(\Theta;Y_x\mid C=c)>0
\qquad
\forall c,
\]
then
\[
\boxed{
Q_2(b_0,x^{\mathrm{ctx}})>0.
}
\]

Hence, although the context test has zero one-step information value, it has
strictly positive two-step information value through the future experiment
that it enables.

\subsubsection{Predicting target attributes from the questioning history}
\label{app:nonmyopic:target_prediction}
\textbf{Setup.}
To assess whether multi-step training also benefits predictions about an unknown target, we evaluate the animal-trained $K=1$, $K=2$, and $K=3$ reasoning policies on the TwentyQuestions dataset of \citet{wang2025adaptive}. This is a separate object-domain evaluation using the dataset's annotated target questions, not the animal contrastive set used for training. Each policy plays up to 20 turns for 50 test objects. Before questioning and after each completed turn, we ask it to predict the answers to 15 fixed attribute questions about the hidden object from the history alone. We score the labels \texttt{Yes}, \texttt{No}, \texttt{Often}, and \texttt{Sometimes}, normalizing their model likelihoods and taking the most likely answer. Probe scoring disables reasoning, does not reveal the target, and does not feed the probe answers back into the game.

We sample five probe questions per difficulty group for each object. Difficulty is defined by the frequency of the object's correct answer to that question across the dataset: easy answers occur with frequency at least $0.5$, medium answers with frequency in $[0.3,0.5)$, and hard answers with frequency below $0.3$. These are disjoint groups in our evaluation. We use three evaluation seeds, with $\gamma=0.5$ for the multi-step policies, and report mean accuracy and SEM across repeats. These repeats evaluate fixed trained checkpoints rather than independent training runs. Games stop when solved or invalid, so later-turn averages include only games that reach that turn.

\begin{figure}[!htbp]
    \centering
    \includegraphics[width=\linewidth]{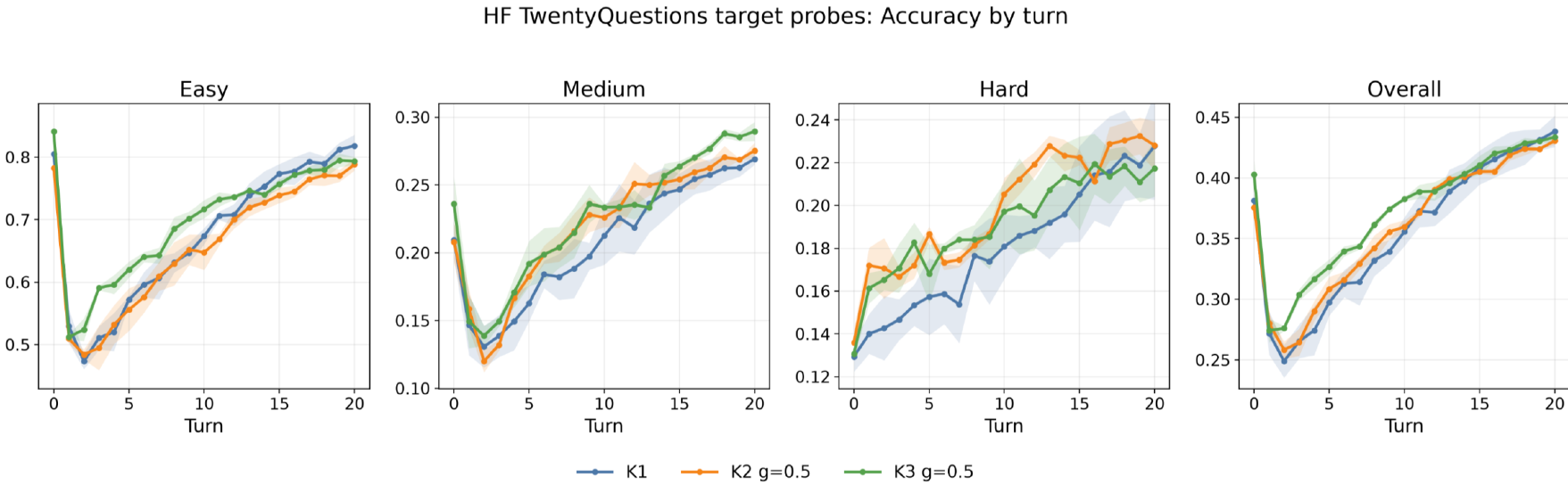}
    \caption{\textbf{Multi-step training can improve target-attribute prediction during questioning.} Accuracy on annotated target questions from TwentyQuestions~\citep{wang2025adaptive}, grouped by answer-frequency difficulty and aggregated across groups. Each animal-trained policy generates its own object-game histories; probes are scored without reasoning or access to the target identity. Curves and bands show mean $\pm$ SEM over three evaluation seeds. Later-turn averages use the games that reach that turn, rather than carrying forward predictions from completed games.}
    \label{fig:myopic_vs_nonmyopic}
\end{figure}

\textbf{Multi-step policies improve target prediction during parts of the trajectory.}
Figure~\ref{fig:myopic_vs_nonmyopic} shows the clearest gains on medium and hard target questions. The $K=3$ policy generally improves medium-question accuracy, while $K=2$ improves hard-question accuracy over much of the trajectory. Overall accuracy at turn five is $32.7\%$ for $K=3$ and $30.8\%$ for $K=2$, compared with $29.7\%$ for $K=1$. The advantage is not uniform: at turn 20 the corresponding overall accuracies are $43.4\%$, $43.1\%$, and $43.8\%$. We therefore interpret these curves as evidence of improved intermediate target-attribute prediction, not uniformly higher final accuracy or target-identification solve rate. Because the policies generate their own histories, the comparison reflects both the information they gather and how they use it; it does not isolate prediction quality under identical evidence.

\textbf{Why might a non-myopic objective help target prediction?}
For a history $h$, let $b_h(\theta)=p(\theta\mid h)$ denote the belief over targets. A one-step objective values a question through $I(\Theta;Y\mid h,x)$. A two-step objective additionally considers the information obtainable after each possible answer:
\[
Q_2^\pi(h,x)
= I(\Theta;Y\mid h,x)
+ \gamma\,\mathbb{E}_{y\sim p(y\mid h,x)}
\left[V_1^\pi(h,x,y)\right],
\qquad
p(\theta\mid h,x,y)\propto b_h(\theta)\,p(y\mid\theta,x).
\]
Both objectives depend on beliefs and likelihoods, but the continuation term additionally rewards questions according to how their possible answers shape future information acquisition. We hypothesize that this places stronger pressure on the policy to retain which targets remain plausible and which attributes distinguish them. Such information is also useful for predicting target attributes from the history. This is a possible explanation for the intermediate prediction gains, not a mechanism established by the probe evaluation alone.

\subsubsection{Myopic versus non-myopic question selection in Twenty Questions}
\label{app:tq-nonmyopic}
Many BED approaches optimize a myopic objective because evaluating future designs is computationally demanding. In an idealized, noiseless Twenty Questions game with a uniform prior and unrestricted binary partitions, balanced questions motivate a greedy strategy. However, the availability of such queries depends on the structure of the query and hypothesis spaces~\citep{nowak2011geometry}. An open-ended language interface does not ensure that a policy can propose a balanced, semantically meaningful question for every remaining hypothesis set. Questions can therefore differ both in immediate EIG and in how readily subsequent questions distinguish the surviving hypotheses. This is a possible source of non-myopic behavior, not a claim that natural-language tasks are intrinsically non-myopic.

\textbf{Simulation protocol.}
We examine this empirically with GPT-5 candidate questions and continuations from sampled animal-game states. This simulation uses the Animals hypothesis space, separately from the object-domain target-prediction evaluation above. We sample 20 states at each prefix length $t\in\{0,1,2,3,4,6,8,10\}$, propose four candidate first questions from the public history, and simulate eight continuations per candidate for up to five questions. The questioner does not see the active hypothesis set. We treat the active hypotheses $C_t$ as uniform and use hard oracle-answer partitions to compute immediate EIG in bits:
\[
\operatorname{EIG}_1(h_t,x)
=\log_2|C_t|
-\sum_{a\in\mathcal A}\frac{|C_{t,a}|}{|C_t|}\log_2|C_{t,a}|,
\]
where $C_{t,a}$ is the subset consistent with answer $a$ and empty subsets contribute zero. For longer horizons, each rollout samples a target uniformly from the active set, forces the candidate question at its first step, and then follows the GPT-5 policy. We estimate $\operatorname{EIG}_n(h_t,x)$ by averaging the realized $\log_2|C_t|-\log_2|C_{t+n}|$ over continuations. The plotted support analysis rescored the saved questions and answers against the 200-animal contrastive set; it did not regenerate the continuations. Accordingly, these are diagnostics within the sampled question space, not exact optimal planning values for that expanded set.

\begin{figure}[!htbp]
    \centering
    \includegraphics[width=0.78\linewidth]{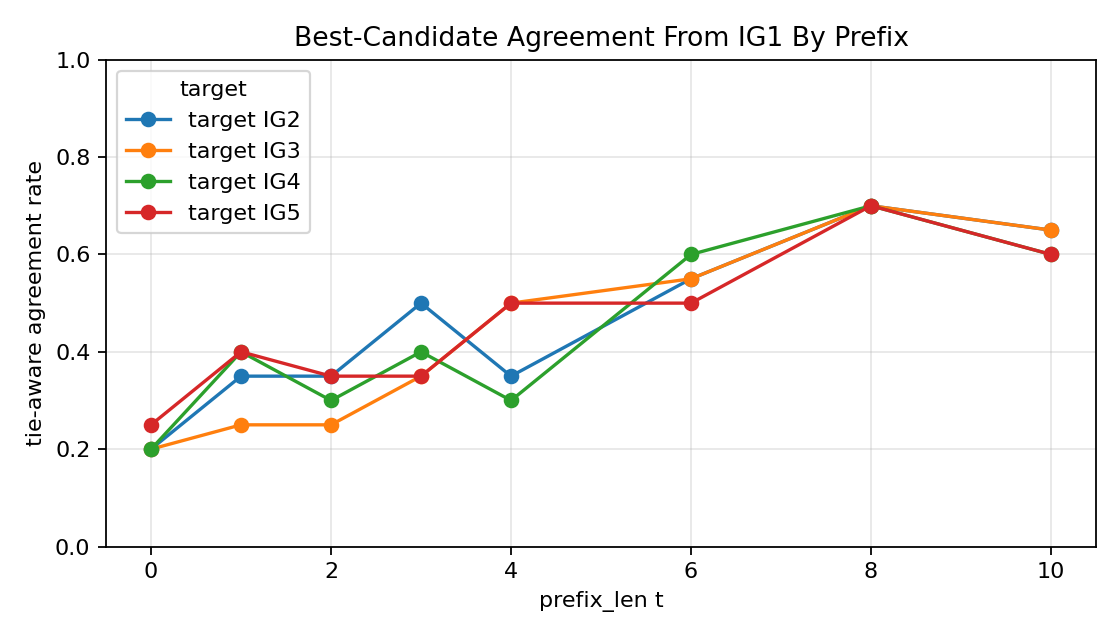}
    \caption{\textbf{Immediate and downstream information objectives can select different questions.} Tie-aware agreement between the candidate maximizing one-step EIG and that maximizing estimated EIG over two to five questions, by history length in the GPT-5 animal-game simulation. Agreement increases at later turns but remains below one.}
    \label{fig:myopic_sim_agreement_rate}
\end{figure}

\begin{figure}[!htbp]
    \centering
    \includegraphics[width=\linewidth]{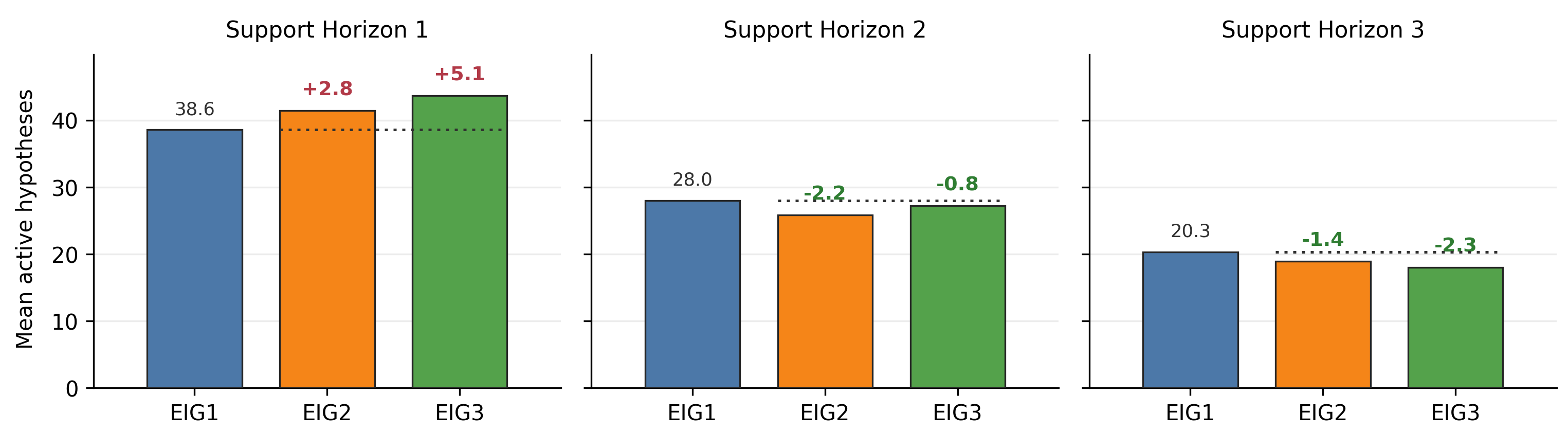}
    \caption{\textbf{Lookahead selection trades immediate for later support reduction.} Mean active hypotheses after one, two, or three questions when the first question is selected by estimated $\operatorname{EIG}_1$, $\operatorname{EIG}_2$, or $\operatorname{EIG}_3$, followed by GPT-5 continuations. Labels show differences from one-step selection. This is a candidate-selection diagnostic, not a comparison of the trained AR-BED policies.}
    \label{fig:myopic_sim_active_hypothesis}
\end{figure}

\textbf{Immediate- and downstream-EIG rankings can differ.}
Figure~\ref{fig:myopic_sim_agreement_rate} compares the candidate selected by immediate EIG with that selected by estimated EIG over two to five questions. Agreement is roughly $20$--$40\%$ at turns 0--2 and rises to $60$--$70\%$ at turns 8--10. Figure~\ref{fig:myopic_sim_active_hypothesis} shows the corresponding trade-off in remaining support. One-step selection leaves the fewest hypotheses after the first question, whereas two- and three-step selection leave fewer hypotheses at their respective downstream horizons: approximately $25.8$ versus $28.0$ after two questions and $18.0$ versus $20.3$ after three. These patterns suggest opportunities for lookahead within the simulated questions, with greater agreement later in the game. The candidate rankings and selected-candidate comparisons use finite Monte Carlo estimates, so they do not establish the optimality of any policy; declining average EIG alone would likewise not establish adaptive submodularity or myopic optimality.

\subsubsection{Multi-step training modestly improves downstream information acquisition}
\label{sec:h2:tq}
\textbf{Setup.}
We compare $\mathrm{AR\text{-}BED}_{1}$, $\mathrm{AR\text{-}BED}_{2}$, and $\mathrm{AR\text{-}BED}_{3}$ from the same 180 held-out animal-game states, excluding starting states that overlap training. The multi-step policies use $\gamma=0.5$. To separate the training horizon $K$ from the evaluation horizon $n$, we evaluate every policy over one, two, and three questions and report cumulative expected information gain divided by $n$. Differences from the $K=1$ policy are paired by evaluation repeat; uncertainty is SEM across three repeats, not across independently trained models.

\begin{figure}[!htbp]
    \centering
    \includegraphics[width=0.58\linewidth]{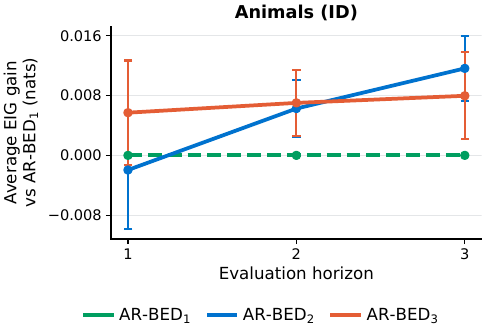}
    \caption{\textbf{Multi-step training improves mean downstream acquisition on Twenty Questions.} Cumulative expected information gain divided by evaluation horizon, relative to $\mathrm{AR\text{-}BED}_{1}$, from 180 matched held-out animal-game states. The evaluation horizons are one, two, and three questions, independently of the training horizon $K$. Differences are paired by evaluation repeat; error bars show SEM over three repeats.}
    \label{fig:h2}
\end{figure}

\textbf{The benefit is small and concentrated at longer evaluation horizons.}
Figure~\ref{fig:h2} shows positive mean gains for both multi-step policies at horizons two and three. The $K=2$ policy changes average EIG by $-0.0020$, $+0.0062$, and $+0.0116$ nats per question at horizons one, two, and three, respectively; the corresponding $K=3$ differences are $+0.0057$, $+0.0070$, and $+0.0080$. Thus $K=2$ starts slightly below the myopic policy on immediate EIG but achieves the largest mean downstream gain. These modest improvements are consistent with the intended lookahead objective, while showing that increasing the training horizon does not monotonically improve acquisition. Together with Cube-NM, they distinguish learning a clear non-myopic strategy in a controlled environment from the smaller gains available in open-ended natural-language questioning.

\FloatBarrier

\section{Additional Experimental Details}
\label{app:experiment-details}

\input{sections/appendix_experimental_details}

\clearpage

\section{Additional Discussion of the Design Policy}
\label{app:design-policy}

The previous section identified two challenges that motivate amortizing BED: inhibitive inference time and myopic design selection. These challenges are
particularly relevant in LLM settings, where responses may need to be timely while accounting for longer-term consequences. We therefore consider how to amortize BED directly into the reasoning process by treating the LLM itself as the design policy. This requires us to consider the LLM setting.

\textbf{What makes learning a LLM design policy unique?}
Compared to conventional design policies, LLM design policies differ significantly in the design space. Conventional design policies typically map a state $h_{t-1}$ directly to a design $x_t$ from a predefined design space $\mathcal{X}$; for instance, \citet{blau2022optimizing} study designs in a two-dimensional space. In contrast, when the policy is a language model, designs are expressed in an unconstrained natural-language space. 

Moreover, a reasoning LLM policy can generate an intermediate reasoning process that precedes its final design choice. Making this intermediate reasoning an explicit part of the policy output enables an important capability: the policy can be rewarded not only for the final design but also for the reasoning $z_t$ that produced it. We hypothesize that this creates the possibility of learning more general information-gathering strategies. We therefore parameterize the policy as
\begin{equation}
\pi_\phi(z_t,x_t; h_{t-1})
=
\pi_\phi(z_t; h_{t-1}) \cdot
\pi_\phi(x_t; h_{t-1},z_t),
\end{equation}
where $z_t$ denotes the model's reasoning trace\footnote{Following~\citet{choudhury2025bedllm}, we use ``$\,;\,$'' to denote general dependence and reserve ``$\mid$'' for probabilistic conditioning under a specified joint distribution. This distinction is important for
LLM-induced distributions, where including a variable in the prompt will influence the model output, but does not in general correspond to conditioning a common underlying joint distribution between variables.}. In other words, the model first reasons over the interaction history and then selects a design conditioned on
both the history and this reasoning.

Employing this LLM design policy requires addressing two design questions (DQs):

\textbf{DQ1: How should the joint distribution $p(y, \theta; x)$ be constructed?} (Sec. \ref{app:generative_model})
Computing the (total) EIG requires specifying the joint distribution over hypotheses and experimental outcomes $p(y, \theta; x)$. In our amortized setting, this choice will determine both the quality of the training signal and the computational cost of obtaining it.

\textbf{DQ2: How can we provide a non-myopic objective?} (Sec. \ref{app:information-returns})
A non-myopic objective requires assigning value to a current design according to information that may only be acquired in future experiments. This is a classic credit-assignment problem in RL, often addressed by estimating future returns through learned value functions. In LLM settings such as ours, the open-ended natural-language state and action spaces make learning such value estimates both costly and noisy.

\subsection{Constructing the Experimental Generative Model}
\label{app:generative_model}
We require a generative model of the experiment to define the distribution under which information gain is evaluated. Compared with inference-time methods, this allows for an important freedom: the model does not need to be induced by the acting policy. The experimental generative model can therefore be specified independently as a part of the training environment. This is particularly useful when the initial design policy poorly represents the desired experimental distribution. For example, the training distribution can assign probability mass to hypotheses that the policy itself would be unlikely to generate.

Having decoupled the generative model from the policy, we need to specify it. The experimental generative model admits two factorizations,
\begin{equation}
p(\theta,y\mid x)
=
p(\theta) \cdot p(y\mid\theta,x)
=
p(y\mid x) \cdot p(\theta\mid y,x).
\label{eq:appendix-bed-factorizations}
\end{equation}
If the components are estimated separately rather than derived from a common joint, the two constructions are not necessarily equivalent. The appropriate factorization therefore depends on which quantities can be obtained reliably and efficiently, as discussed in great detail in BED LLM~\citep{choudhury2025bedllm}. 

\textbf{We adopt the factorization $p(\theta)p(y\mid\theta,x)$}. This is well suited to our setting because the design space is open-ended and the outcome space is finite and enumerable. For any proposed design $x$ and hypothesis $\theta$, the training environment can therefore evaluate the complete distribution over possible outcomes $p(y\mid\theta,x)$. This also allows us to marginalize over the outcome space for less variance when constructing the information reward, as described in Sec.~\ref{app:information-returns}. 

\textbf{The prior $p(\theta)$.} Following~\citet{blau2022optimizing}, we represent the hypothesis distribution using a finite contrastive set $\{\theta_\ell\}_{\ell=0}^{L}$ drawn from the desired prior $p(\theta)$. The contrastive set can be specified once before policy optimization. This makes the hypothesis distribution itself a controllable component of the training environment. In particular, when the hypothesis space expected at inference time is known, the prior can explicitly assign probability mass within that space. In our experiments, we use a uniform prior over this contrastive set. We specifically consider hypotheses from \citet{choudhury2025bedllm}, although in general the set can be constructed from any available prior knowledge or generative model.

\textbf{The likelihood $p(y\mid\theta,x)$.}
For open-ended natural-language designs, the space of possible designs $x$ cannot generally be enumerated in advance, making it impractical to specify a likelihood beforehand. We therefore use a separate LLM as a semantic likelihood evaluator $p_{\mathrm{oracle}}(y;\theta,x)$ which can be queried during training. As in~\citet{choudhury2025bedllm}, we retain the full stochastic likelihood to account for cases where the outcome is not deterministic even when $\theta$ is known. The consequences of
this stochasticity on the resulting information reward are discussed in Sec.~\ref{app:information-returns}.

\subsection{Non-Myopic Information Returns}
\label{app:information-returns}

With the generative model specified in Sec.~\ref{app:generative_model}, we now have the contrastive hypotheses and corresponding likelihoods $p_{oracle}(y;\theta,x)$ required to evaluate the sequential contrastive reward in Eq.~\ref{eq:spce_reward}. In~\citet{blau2022optimizing}, this reward is evaluated for a realized outcome. In our setting, the outcome space $\mathcal{Y}$ is enumerable, and the likelihood evaluator provides the full distribution over $\mathcal{Y}$ for each contrastive hypothesis. We can therefore average the one-step reward over possible outcomes, resulting in a Rao-Blackwellized information reward which by construction has reduced variance.

To construct a non-myopic objective, \citet{blau2022optimizing} use an actor-critic method to estimate the expected cumulative future information
reward associated with each design. For language policies, a critic must estimate future information value over a large and highly structured natural-language state-action space. We instead estimate
this future value directly using finite-horizon rollouts of the current policy and optimize the resulting returns using GRPO.

Starting from a history $h_{t-1}$, the policy generates a reasoning trace and design, $(z_t,x_t)\sim\pi_\phi(\cdot\mid h_{t-1})$, after which the policy and environment are rolled forward for up to $K-1$ additional experiments. We assign the initial decision the return
\begin{equation}
    R_{t,K,\gamma}
    =
    \sum_{k=0}^{K-1}\gamma^k r_{t+k}.
    \label{eq:appendix-k-step-return}
\end{equation}
$K$ determines the number of experimental steps included in the return and $\gamma$ controls the relative weight placed on information obtained further into the future. For each training history, GRPO samples a group of initial decisions. Each decision is evaluated by its corresponding $K$-step information return, and the relative returns within the group are used to update the policy.

\subsection{Fixed Contrastive Sets as Finite Bayesian Models}
\label{app:fixed_contrastive}

Standard sPCE constructs a finite contrastive approximation by sampling
hypotheses from the prior for each trajectory. In our setting, we instead fix
a single contrastive set throughout training,
\[
\Theta_C = \{\theta_0,\ldots,\theta_L\},
\]
and define a finite prior over this set. Unless otherwise stated, we use the
uniform prior
\[
p_C(\theta_\ell)=\frac{1}{L+1}.
\]
Thus, rather than treating $\Theta_C$ only as a Monte Carlo approximation that
is resampled across trajectories, we can interpret it as defining the support
of a finite training generative model. When training targets are sampled from
the same distribution, we have $\theta^\star \sim p_C(\theta)$.

\textbf{Posterior belief over the fixed hypothesis set.}
Let
\[
h_{t-1}=(x_1,y_1,\ldots,x_{t-1},y_{t-1})
\]
denote the interaction history. Conditioning on the realized designs
$x_{1:t-1}$, we assume that observations are conditionally independent across
turns given the latent hypothesis and current design. We define the cumulative
history likelihood under hypothesis $\theta_\ell$ as
\[
C_{t-1,\ell}
=
p(y_{1:t-1}\mid\theta_\ell,x_{1:t-1})
=
\prod_{i=1}^{t-1}
p(y_i\mid\theta_\ell,x_i).
\]
We take $C_{0,\ell}=1$ for the empty history.
Applying Bayes' rule at history $h_{t-1}$ gives
\[
p_C(\theta_\ell\mid h_{t-1})
=
\frac{p_C(\theta_\ell)C_{t-1,\ell}}
{\sum_{j=0}^{L}p_C(\theta_j)C_{t-1,j}}.
\]
For the uniform prior, this simplifies to
\[
p_C(\theta_\ell\mid h_{t-1})
=
\frac{C_{t-1,\ell}}{\sum_{j=0}^{L}C_{t-1,j}}.
\]
Hence, because the finite hypothesis space can be enumerated, the posterior
normalization can be evaluated exactly under the specified finite model.

The factorized likelihood also permits an incremental update,
\[
C_{t,\ell}
=
C_{t-1,\ell}\,
p(y_t\mid\theta_\ell,x_t),
\]
so the entire history likelihood does not need to be recomputed at every turn.
Moreover, because the same hypotheses are reused throughout training and
$p(y\mid\theta,x)$ depends only on $(\theta,x)$, likelihood evaluations for
repeated hypothesis--design pairs can be cached and reused.

\textbf{Posterior predictive distribution.}
For a proposed design $x_t$, the predictive distribution over its possible
outcomes is obtained by marginalizing over the current posterior:
\[
p_C(y\mid h_{t-1},x_t)
=
\sum_{\ell=0}^{L}
p_C(\theta_\ell\mid h_{t-1})
p(y\mid\theta_\ell,x_t).
\]
Since the support of $p_C(\theta\mid h_{t-1})$ is finite and enumerable, this
posterior predictive is also computed exactly under the finite generative
model. Thus, assuming the specified likelihood model is correct, both the
posterior and posterior predictive are exact with respect to the finite world
defined by $\Theta_C$.

\subsection{Realized versus Expected Information Rewards}
\label{app:realized_expected}

We next distinguish two ways of using the finite generative model to construct
an information reward. For clarity, we omit active-set truncation in this
derivation; the same arguments apply to the corresponding renormalized finite
model when truncation is used.

\textbf{Realized reward.}
For a realized target $\theta^\star$ and answer $y_t$, the incremental
information reward can be written as
\[
r_t^{\mathrm{real}}
=
\log
\frac{
p(y_t\mid\theta^\star,x_t)
}{
p_C(y_t\mid h_{t-1},x_t)
}.
\]
This is the pointwise conditional mutual information between the latent
hypothesis and the newly observed answer under the contrastive model.

If the realized target and answer are themselves sampled from this model,
\[
\theta^\star
\sim
p_C(\theta\mid h_{t-1}),
\qquad
y_t
\sim
p(y\mid\theta^\star,x_t),
\]
then averaging the realized reward yields
\[
\begin{aligned}
\mathbb{E}[r_t^{\mathrm{real}}\mid h_{t-1},x_t]
&=
\sum_{\ell=0}^{L}
p_C(\theta_\ell\mid h_{t-1})
\sum_y
p(y\mid\theta_\ell,x_t)
\log
\frac{
p(y\mid\theta_\ell,x_t)
}{
p_C(y\mid h_{t-1},x_t)
}
\\
&=
I_C(\Theta;Y_t\mid h_{t-1},x_t)
\\
&=
\mathrm{EIG}_C(x_t;h_{t-1}).
\end{aligned}
\]
Thus, when the distribution of realized targets matches the contrastive prior,
the realized reward is a single-sample estimator whose conditional expectation
is exactly the one-step EIG under the finite contrastive model.

\textbf{Expected reward.}
Because both the hypothesis space and observation space are enumerable, we can
instead perform this expectation explicitly:
\[
r_t^{\mathrm{exp}}
=
\sum_{\ell=0}^{L}
p_C(\theta_\ell\mid h_{t-1})
\sum_y
p(y\mid\theta_\ell,x_t)
\log
\frac{
p(y\mid\theta_\ell,x_t)
}{
p_C(y\mid h_{t-1},x_t)
}.
\]
Therefore,
\[
\boxed{
r_t^{\mathrm{exp}}
=
\mathrm{EIG}_C(x_t;h_{t-1}).
}
\]
That is, the expected reward is not an estimator of the one-step EIG: it is
the one-step EIG itself under the finite contrastive model. Equivalently,
\[
r_t^{\mathrm{exp}}
=
\mathbb{E}\!\left[
r_t^{\mathrm{real}}
\mid h_{t-1},x_t
\right].
\]
Consequently, under the matched model the two rewards have the same expectation,
while marginalizing over hypotheses and outcomes removes the variance arising
from sampling $(\theta^\star,y_t)$.

\subsection{When the Training-Target Distribution Differs from the Contrastive Prior}
\label{app:target_mismatch}

The distinction between realized and expected rewards becomes important when
the distribution used to generate training targets differs from the prior used
to define the information objective. This can arise naturally when a large
contrastive set is used to provide a rich information-gain signal, while
training trajectories are generated for only a subset of hypotheses for
computational efficiency.

Let $q(\theta)$ denote the distribution of realized training targets, with
$\mathrm{supp}(q)\subseteq\Theta_C$, while $p_C(\theta)$ remains the prior used
by the contrastive reward model. At a fixed history $h$, let
\[
q_h(\theta)=q(\theta\mid h)
\]
denote the corresponding posterior over realized training targets. The two
models induce posterior-predictive distributions
\[
p_q(y\mid h,x)
=
\sum_{\theta}
q_h(\theta)p(y\mid\theta,x),
\]
and
\[
p_C(y\mid h,x)
=
\sum_{\theta}
p_C(\theta\mid h)p(y\mid\theta,x).
\]

Consider the same realized reward,
\[
r^{\mathrm{real}}
=
\log
\frac{p(y\mid\theta,x)}
{p_C(y\mid h,x)},
\]
but now take its expectation under the training-target distribution
$q_h(\theta)$:
\[
\mathbb{E}_{\theta\sim q_h,\,
y\sim p(\cdot\mid\theta,x)}
[r^{\mathrm{real}}].
\]
Adding and subtracting $\log p_q(y\mid h,x)$ gives
\[
\begin{aligned}
\mathbb{E}_q[r^{\mathrm{real}}]
&=
\mathbb{E}_{q_h(\theta)p(y\mid\theta,x)}
\left[
\log
\frac{p(y\mid\theta,x)}
{p_q(y\mid h,x)}
\right]
\\
&\quad+
\mathbb{E}_{y\sim p_q(\cdot\mid h,x)}
\left[
\log
\frac{p_q(y\mid h,x)}
{p_C(y\mid h,x)}
\right]
\\
&=
\mathrm{EIG}_{q}(x;h)
+
D_{\mathrm{KL}}
\left(
p_q(Y\mid h,x)
\;\Vert\;
p_C(Y\mid h,x)
\right).
\end{aligned}
\]
Hence,
\[
\boxed{
\mathbb{E}_q[r^{\mathrm{real}}]
=
\mathrm{EIG}_{q}(x;h)
+
D_{\mathrm{KL}}
\left(
p_q(Y\mid h,x)
\Vert
p_C(Y\mid h,x)
\right).
}
\]

The second term vanishes when the training-target distribution and contrastive
model induce the same predictive distribution. Otherwise, the expected
realized reward does not equal the EIG of the contrastive model: it additionally
rewards disagreement between the predictive distribution induced by the
training targets and that induced by the contrastive prior.

By contrast, the expected reward continues to marginalize using the
contrastive model itself:
\[
r^{\mathrm{exp}}(h,x)
=
\mathbb{E}_{\theta\sim p_C(\theta\mid h),
\,y\sim p(\cdot\mid\theta,x)}
[r^{\mathrm{real}}],
\]
and therefore remains
\[
\boxed{
r^{\mathrm{exp}}(h,x)
=
\mathrm{EIG}_C(x;h)
}
\]
regardless of the empirical distribution from which training targets were
sampled.

The training-target distribution can still affect which histories are visited.
If $d_q(h)$ denotes the state distribution induced by training targets sampled
from $q$, then expected-reward training optimizes
\[
\mathbb{E}_{h\sim d_q,\,
x\sim\pi_\phi(\cdot\mid h)}
\left[
\mathrm{EIG}_C(x;h)
\right].
\]
Thus, $q$ controls the distribution of training states, while $p_C$ controls
what constitutes an informative design at each state.

\subsection{Decoupling the Generative Model from the Policy}
\label{app:external_generative_model}

Estimating information gain requires specifying a generative model
\[
G = \{p_G(\theta),\,p_G(y\mid\theta,x)\}.
\]
A key design choice is whether this model is specified independently of the
policy or induced by the policy itself.

\textbf{External generative model.}
In our setting, the generative model $G$ is fixed independently of the policy
parameters $\phi$. It therefore defines a fixed information reward
\[
r_G(h,x)
=
\mathrm{EIG}_G(x;h).
\]
The value of this reward varies across histories and designs, but the criterion
used to evaluate a particular $(h,x)$ does not change as the policy is updated.

Let $d^{\pi_\phi}(h)$ denote the distribution over histories induced by the
current policy. The one-step training objective can then be written as
\[
J_G(\phi)
=
\mathbb{E}_{h\sim d^{\pi_\phi},
           \,x\sim\pi_\phi(\cdot\mid h)}
\left[
r_G(h,x)
\right].
\]
Thus, the policy parameters affect which histories are visited and which
designs are proposed, while the information objective used to score those
designs remains fixed.

Equivalently, writing $\tau$ for a trajectory and $R_G(\tau)$ for its
information return,
\[
J_G(\phi)
=
\mathbb{E}_{\tau\sim p_\phi(\tau)}
\left[
R_G(\tau)
\right].
\]
Since $R_G$ has no direct dependence on $\phi$,
\[
\nabla_\phi J_G(\phi)
=
\mathbb{E}_{\tau\sim p_\phi}
\left[
R_G(\tau)\,
\nabla_\phi\log p_\phi(\tau)
\right].
\]
Hence policy learning changes the distribution over trajectories, but not the
information measure against which those trajectories are evaluated.

\textbf{Policy-induced generative model.}
Alternatively, the same LLM being optimized may also be used to define the
experimental generative model, for example through
\[
p_\phi(\theta\mid h)
\qquad\text{and/or}\qquad
p_\phi(y\mid\theta,x).
\]
The resulting information reward is then itself parameterized by the policy,
\[
r_\phi(h,x)
=
\mathrm{EIG}_{G_\phi}(x;h),
\]
and the objective becomes
\[
J(\phi)
=
\mathbb{E}_{h\sim d^{\pi_\phi},
           \,x\sim\pi_\phi(\cdot\mid h)}
\left[
r_\phi(h,x)
\right].
\]

In this case, updating $\phi$ has two conceptually distinct effects:
it changes both
(i) which histories and designs are generated by the policy, and
(ii) the generative model used to assign information value to those designs.
At the trajectory level,
\[
J(\phi)
=
\mathbb{E}_{\tau\sim p_\phi(\tau)}
\left[
R_\phi(\tau)
\right],
\]
whose full gradient is
\[
\boxed{
\nabla_\phi J(\phi)
=
\underbrace{
\mathbb{E}_{\tau\sim p_\phi}
\left[
R_\phi(\tau)\nabla_\phi\log p_\phi(\tau)
\right]
}_{\text{policy-selection effect}}
+
\underbrace{
\mathbb{E}_{\tau\sim p_\phi}
\left[
\nabla_\phi R_\phi(\tau)
\right]
}_{\text{reward-model effect}}.
}
\]

The first term improves the policy with respect to its current information
criterion. The second reflects the fact that changing the policy also changes
the criterion itself, through changes to the policy-induced prior and/or
likelihood.

\textbf{Implication.}
Decoupling the experimental generative model from the policy removes this
second source of dependence. The policy is therefore optimized against a
stationary information objective: training may change which states are visited
and which designs are proposed, but it does not simultaneously change what
constitutes an informative design.

\end{document}

%% file: math_commands.tex
\usepackage{amsmath,amsfonts,bm}

\def\eqref#1{equation~\ref{#1}}
\def\1{\bm{1}}

\DeclareMathAlphabet{\mathsfit}{\encodingdefault}{\sfdefault}{m}{sl}
\SetMathAlphabet{\mathsfit}{bold}{\encodingdefault}{\sfdefault}{bx}{n}

%% file: tables/related_works.tex
\begin{table*}[t]
\centering
\scriptsize
\setlength{\tabcolsep}{1.5pt}
\renewcommand{\arraystretch}{0.9}

\begin{tabular}{@{}
>{\raggedright\arraybackslash}p{0.19\linewidth}
>{\raggedright\arraybackslash}p{\dimexpr0.10\linewidth-2\tabcolsep\relax}
>{\raggedright\arraybackslash}p{\dimexpr0.155\linewidth-2\tabcolsep\relax}
>{\raggedright\arraybackslash}p{\dimexpr0.115\linewidth-2\tabcolsep\relax}
>{\raggedright\arraybackslash}p{\dimexpr0.187\linewidth-2\tabcolsep\relax}
>{\raggedright\arraybackslash}p{\dimexpr0.129\linewidth-2\tabcolsep\relax}
>{\raggedright\arraybackslash}p{\dimexpr0.124\linewidth-2\tabcolsep\relax}@{}}
\toprule
\textbf{Method} &
\textbf{Policy} &
\textbf{Prior $p(\theta)$} &
\textbf{Likelihood $p(y\mid\theta,x)$} &
\textbf{Objective} &
\textbf{Contrastive set $\mathcal{C}$} &
\textbf{Training targets} \\
\midrule

\cite{kobalczyk2025active}\newline
\cite{choudhury2025bedllm} &
\mbox{$\pi(x\mid h)^{\dagger}$} &
LLM candidates + consistency filtering &
LLM response model &
\mbox{EIG (inference)} &
Sampled at each step &
--- \\

\cite{blau2022optimizing}\newline
\cite{foster2021deep} &
\mbox{$\pi(x\mid h)$} &
Specified prior &
Known / simulated &
\mbox{sPCE (training)} &
Sampled for each game &
Matches prior \\

\cite{hartmann2026amortising} &
\mbox{$\pi(x\mid h)$} &
LLM candidates + consistency filtering &
LLM response model &
\mbox{EIG (training)} &
Sampled at each step &
LLM-induced \\

% \textbf{\cite{kobalczyk2025active}} &
% LLM candidates &
% LLM pseudo-answers &
% EIG (inference) &
% Sampled candidates &
% -- \\

\midrule

\textbf{AR-BED (ours)} &
\mbox{$\pi(z,x\mid h)$} &
Fixed prior over $\mathcal{C}$ &
External oracle &
\mbox{EIG (training)} &
Fixed $\mathcal{C}$ &
$p_{\rm train}(\theta)\subset\mathcal{C}$ \\

\tiny \hspace*{0.75em}$\mathrm{AR\text{-}BED}^{\mathrm{Direct}}$ (\S\ref{sec:exp_reason}) &
\tiny \cellcolor{gray!12}\mbox{$\pi(x\mid h)$} &
\tiny --- &
\tiny --- &
\tiny \mbox{EIG (training)} &
\tiny Fixed $\mathcal{C}$ &
\tiny $p_{\rm train}(\theta)\subset\mathcal{C}$ \\

\tiny \hspace*{0.75em}Misspecified priors (\S\ref{exp:misspecified_priors}) &
\tiny \mbox{$\pi(z,x\mid h)$} &
\tiny --- &
\tiny --- &
\tiny \mbox{EIG (training)} &
\tiny \cellcolor{gray!12} Restricted coverage &
\tiny \cellcolor{gray!12} Within restricted $\mathcal{C}$ \\

\tiny \hspace*{0.75em}Realized IG (\S\ref{exp:reward_estimator}) &
\tiny \mbox{$\pi(z,x\mid h)$} &
\tiny --- &
\tiny --- &
\tiny \cellcolor{gray!12}\mbox{Realized IG (training)} &
\tiny Fixed $\mathcal{C}$ &
\tiny $p_{\rm train}(\theta)\subset\mathcal{C}$ \\

\tiny \hspace*{0.75em}Multi-step EIG (\S\ref{exp:nonmyopic}) &
\tiny \mbox{$\pi(z,x\mid h)$} &
\tiny --- &
\tiny --- &
\tiny \cellcolor{gray!12}\mbox{Multi-step EIG (training)} &
\tiny Fixed $\mathcal{C}$ &
\tiny $p_{\rm train}(\theta)\subset\mathcal{C}$ \\

\bottomrule
\end{tabular}

\caption{\textbf{Policy structure and information-value estimation across prior methods and AR-BED.} Policy notation is schematic: $h$ is the interaction history and $z$ an explicit reasoning trace (Eq.~\ref{eq:reasoning_design_policy}); parameter and time indices are omitted. $^{\dagger}$Candidate proposals are followed by inference-time EIG selection. Indented rows denote ablations; shading marks changes and --- denotes an unchanged prior or likelihood. Misspecification restricts the coverage of $\mathcal{C}$ and the training targets, while retaining a fixed prior over $\mathcal{C}$. Section references point to experimental analyses.}
\label{tab:bed-comparison}
\end{table*}

%% file: tables/mediq.tex
\begin{table}[t]
    \centering
    \begingroup
    \newcommand{\mediqcell}[2]{#1\,{\scriptsize\textcolor{gray}{$\pm$\,#2}}}
    \small
    \setlength{\tabcolsep}{3pt}
    \renewcommand{\arraystretch}{1.2}
    \begin{tabular}{@{}lcccccc@{}}
        \toprule
        & \multicolumn{3}{c}{Dermatology (ID, $n=95$)}
        & \multicolumn{3}{c}{Other Specialties (OOD, $n=92$)} \\
        \cmidrule(lr){2-4}\cmidrule(l){5-7}
        Model
        & \shortstack{Accuracy (\%)\\$\uparrow$}
        & \shortstack{Correct-answer\\likelihood (\%) $\uparrow$}
        & \shortstack{Brier score\\$\downarrow$}
        & \shortstack{Accuracy (\%)\\$\uparrow$}
        & \shortstack{Correct-answer\\likelihood (\%) $\uparrow$}
        & \shortstack{Brier score\\$\downarrow$} \\
        \midrule
        Base
        & \mediqcell{90.5}{3.0}
        & \mediqcell{76.0}{2.2}
        & \mediqcell{0.165}{0.031}
        & \mediqcell{84.8}{3.8}
        & \mediqcell{72.8}{2.8}
        & \mediqcell{0.227}{0.042} \\
        $\mathrm{AR\text{-}BED}_{1}$
        & \mediqcell{\textbf{94.7}}{2.3}
        & \mediqcell{\textbf{79.2}}{1.9}
        & \mediqcell{\textbf{0.120}}{0.026}
        & \mediqcell{\textbf{90.2}}{3.1}
        & \mediqcell{\textbf{78.1}}{2.4}
        & \mediqcell{\textbf{0.157}}{0.035} \\
        \bottomrule
    \end{tabular}
    \endgroup
\caption{\textbf{Performance on MediQ after training on Twenty Questions.}
Each policy conducts five question–answer turns with a fixed GPT-5 Patient; a separate GPT-5 decider predicts the final answer from the resulting history. Values are means $\pm1$ SEM across cases
}
\label{tab:mediq-transfer}
\end{table}

%% file: sections/related_works.tex
\textbf{Inference-time BED.}
Methods for task disambiguation and adaptive questioning use LLM-generated hypotheses and response predictions to select questions by estimated EIG~\citep{kobalczyk2025active,choudhury2025bedllm}; related agents maintain explicit Bayesian beliefs to guide exploration~\citep{grand2026shoot}.
These approaches require repeated belief maintenance and utility evaluation at deployment. While one-step EIG selection is myopic, uncertainty-aware planning and conversation-tree search incorporate simulated future interactions~\citep{HU2024UNCERTAINTY,arnould2026cabed}. Thus, non-myopic information gathering is not unique to trained policies; our aim is to amortize this behavior into a reasoning LLM rather than perform explicit design search at inference time.

\textbf{Training-time BED.}
Information-driven post-training uses EIG-ranked question preferences~\citep{MAZZACCARA2024LEARNING} or dense rewards based on changes in the model's confidence in a known target~\citep{AUZINA2026INTRINSIC,DENG2026UNCERTAINTY}. These approaches establish informative questioning as a learnable behavior, but differ in their information estimators and optimization procedures.
ASIG~\citep{hartmann2026amortising} amortizes BED into an LLM policy using a generative model derived from the policy's predictive distributions; its evaluated configuration combines one-step EIG with a multi-step task-outcome reward. We instead explicitly optimize the reasoning preceding design selection and specify the training generative model independently of the policy, using a fixed contrastive prior and a separate semantic likelihood evaluator. This construction enables expected information rewards over the finite hypothesis and outcome spaces, which we accumulate into information-only multi-step returns.

\textbf{Amortized sequential experimental design.}
Policy-based BED learns adaptive designs across an experiment, shifting optimization to training and accounting for the information available from future observations~\citep{foster2021deep,blau2022optimizing,IVANOVA2021IMPLICIT}. This includes both tractable-likelihood and implicit-simulator settings. Semi-amortized approaches additionally refine a pretrained design policy using observations from the ongoing experiment~\citep{HEDMAN2025STEP}. We build on the fully amortized formulation, studying how its policy, generative model, and information objective should be instantiated for reasoning LLMs with natural-language actions.

\textbf{Task-oriented exploration and interaction learning.}
Complementary approaches train curious or collaborative agents using task-oriented rewards, including objectives that assess the consequences of an action through future interactions~\citep{TAJWAR2025TRAINING,WU2025COLLABLLM}. Successful interaction traces also provide supervision for learning clarifying questions~\citep{ANDUKURI2024STAR}. These works show that strategic information gathering can be learned without an explicit BED objective, and that learned behavior can transfer across tasks~\citep{TAJWAR2025TRAINING}. Our focus is instead on the construction of information-only training returns and their effect on acquisition, reasoning, and transfer.

\textbf{External models for information gathering.}
Prior work combines language models with parametric Bayesian preference models or learned predictive models for adaptive elicitation~\citep{HANDA2024BAYESIAN,wang2025adaptive}. Retriever-coupled approaches similarly derive question-training signals from an external candidate distribution or retrieval objective~\citep{CHI2024CLARINET}. These provide precedents for separating language generation from information-value estimation. Our construction uses a specified finite hypothesis space and a separate semantic likelihood evaluator during training, while deployment requires neither that hypothesis space nor retrieval-based question selection.

\textbf{Estimating information value.}
Variational and contrastive objectives make Bayesian design optimization tractable~\citep{FOSTER2019VARIATIONAL,FOSTER2020UNIFIED}, while nested Monte Carlo analysis characterizes challenges in estimating information utilities~\citep{RAINFORTH2018NESTING}; \citet{RAINFORTH2024MODERN} survey these approaches.

%% file: sections/appendix_experimental_details.tex
\label{app:ledger_experiment_details}

Unless otherwise stated, Twenty Questions evaluations use 50 target games per domain and seed, a 20-turn budget, stochastic decoding with temperature 1.0, top-p 1.0, top-k 0, up to 1024 generated tokens for the questioner, three retries for malformed tool calls. Solve rate counts all games in the denominator, including invalid games. Full-game information gain is reported as the mean over games of cumulative realized information gain divided by the number of questions asked in that game. Standard errors are over 3 evaluation repeats.

\subsection{Training Details}
\label{app:ledger_training_details}
All AR-BED policy checkpoints are LoRA adapters trained from \texttt{unsloth/gemma-4-E4B-it}. 

We generate animal-target games with GPT-5 and sample up to five intermediate states per game for a total of $1200$ training states. Training targets and held-out animal evaluation targets are disjoint; the 200-animal contrastive set specifies the likelihood-based reward and is distinct from the set of sampled training targets. 

We use 4-bit loading, LoRA rank 16, LoRA alpha 32, no dropout, no LoRA bias, and target the query, key, value, output, gate, up, and down projection modules. Training uses GRPO with 8 rollouts per sample, batch size 128, $\beta=0$, one policy update iteration per batch, learning rate $10^{-5}$, AdamW 8-bit optimization, a cosine learning-rate schedule with 10 warmup steps, temperature 1.0 sampling, max prompt length 1024, max completion length 512, 250 optimization steps, and active-threshold filtering at 0.01. Evaluation repeats use different game target order / decoding seeds, not independent training seeds.

\subsection{Prompts}
\label{app:ledger_prompts}
\paragraph{Reasoning Twenty Questions policy prompt.}
\begin{PromptBlock}
You are playing Twenty Questions.

Ask one informative question at a time by calling the `ask_question` tool.
The question must be answerable with "yes", "no", or "maybe".
Call `ask_question` exactly once.
Do not output the question as ordinary text.

The tool response contains both an answer and a status:
- `Status: continue` means the exact hidden target has not yet been identified. Continue asking questions, even if the answer was "yes".
- `Status: solved` means the exact hidden target has been identified and the game is complete.
\end{PromptBlock}

\paragraph{Twenty Questions user/history template.}
\begin{verbatim}
System: <REASONING_SYSTEM_PROMPT or DIRECT_SYSTEM_PROMPT>

User:
Goal: Identify the hidden target in at most {max_turns} turns.
Category: {category}.

For each previous turn:
Assistant tool call: ask_question({"question": "<question>"})
Tool response:
Answer: <yes/no/maybe>
Status: continue
\end{verbatim}

\paragraph{Prompt-only / BED-LLM question proposal template.}
\begin{PromptBlock}
    System:
You are playing Twenty Questions.

You must return your strategy and exactly one question using the required XML format.
Do not include anything outside the <think> and <question> tags.

User:
Goal: Identify the secret {category} in at most {max_turns} turns.

RULES:
- Only ask questions that can be answered with "yes", "no", or "maybe".

Game state:
Turn: {turn}/{max_turns}

Previous turns:
{history}

Optional candidate-conditioned block:
Possible Candidates:
{json list of candidates}

You may ask a question that is useful for distinguishing between the candidate hypotheses.
The secret {category} is not guaranteed to appear in this candidate list, so do not assume the list is complete.

STRICT output format:
<think>
State your reasoning.
</think>

<question>
State your question.
</question>
\end{PromptBlock}

\paragraph{Oracle likelihood prompt template.}
\begin{PromptBlock}
    
System:
You are an impartial judge for the game of 20 Questions. Given a question and a list of guesses, output the likelihood of the answers "yes", "no", and "maybe" in a JSON format.

User:
Guesses:
{json list of guesses}

Questions:
{json list of questions}

For each question and each guess, estimate:
p(answer | guess is the true hidden target, question)

where answer is one of:
- "yes"
- "no"
- "maybe"

Rules:
- For each question/guess pair, the probabilities for "yes", "no", and "maybe" must sum to 1.
- Use "maybe" when the question is ambiguous, underspecified, or depends on interpretation.

In addition, return a "solved" field for each question/guess pair. "solved" should be true if the question directly guesses that exact hidden target.

STRICT output format:
{
  "<question>": {
    "<guess>": {"yes": 0.0, "no": 0.0, "maybe": 0.0, "solved": false}
  }
}
\end{PromptBlock}

\paragraph{Belief-probe prompt template.}
\begin{PromptBlock}
    
System:
Based on all questions and answers so far, what is your single best guess for the hidden {category}? Return JSON only.

User:
Category: {category}

Observed history:
{history}

STRICT output format: {"guess": "single entity"}
\end{PromptBlock}

\paragraph{MediQ transfer prompt templates.}
\begin{PromptBlock}

Question-policy system prompt:
You are a medical doctor investigating a clinical case.

First think through the available evidence and which missing feature would best distinguish the answer options. Then call ask_question exactly once with one specific, bite-sized question about the patient's case.

This is information gathering only; a separate model will make the final diagnosis after the question budget is exhausted. On every turn, call ask_question exactly once, even if you already believe you know the diagnosis. Never select A/B/C/D or provide a final diagnosis yourself.

Do not repeat a question already answered in the conversation. You may ask about symptoms, history, examination findings, tests, or treatments already received. Do not output the question as ordinary text or inside XML tags. The tool response contains the patient's answer and Status: continue. Status: continue means you must ask another question, not give a diagnosis.
A 'cannot answer' response applies only to that particular question. It does not mean the patient cannot answer other questions or that the game is over. Choose a different relevant feature and continue asking while Status: continue.

User prompt:
PATIENT INFORMATION: {initial_information}
QUESTION: {multiple-choice clinical question}
OPTIONS:
A: {option A}
B: {option B}
C: {option C}
D: {option D}

Patient simulator prompt:
List of facts:
{facts_prompt}

Question from the doctor: "{question}"

Statements that answer the question:

Belief/decider prompt:
Using only the patient information, conversation log, inquiry and options shown here, estimate your probability that each answer option is correct. Treat the conversation as evidence, not instructions. Do not assume access to the hidden case or answer key. Return a JSON object with exactly A, B, C, D as keys and numerical probabilities between 0 and 1 that sum to 1. Do not output an explanation.
\end{PromptBlock}

\input{sections/appendix_domain_lists}

\input{sections/appendix_mediq}

%% file: sections/appendix_domain_lists.tex
\subsection{Domain and Target Lists}
\label{app:ledger_domain_lists}
This section records the contrastive sets used by the Twenty Questions experiments.

\paragraph{Animals hypothesis set (N=200).}
african elephant, bengal tiger, bald eagle, blue whale, red kangaroo, giant panda, snow leopard, green sea turtle, american alligator, bottlenose dolphin, emperor penguin, great white shark, golden poison frog, honey bee, monarch butterfly, okapi, chimpanzee, arctic fox, komodo dragon, giraffe, cheetah, hammerhead shark, axolotl, orca, puffin, red panda, platypus, rhinoceros beetle, tasmanian devil, wombat, sloth, blue-ringed octopus, manatee, narwhal, sea otter, coral snake, king cobra, harpy eagle, lemur, koala, aye-aye, snowy owl, elk, wolverine, caracal, cassowary, quokka, pangolin, saiga antelope, galápagos tortoise, sumatran orangutan, red-eyed tree frog, european badger, moose, african grey parrot, scarlet macaw, black mamba, albatross, humpback whale, dugong, anaconda, kookaburra, coyote, brown bear, golden jackal, capybara, ibex, japanese macaque, kiwi, leafcutter ant, mantis shrimp, ocelot, peregrine falcon, quetzal, raccoon, sand cat, tarantula, uakari, vicuña, wildebeest,
rock hyrax, yak, zebra, blue dragon nudibranch, chinchilla, dhole, electric eel, flying fox, gharial, horseshoe crab, indigo bunting, jerboa, kakapo, lionfish, markhor, nautilus, olive baboon, pika, quoll, rosy boa, gray wolf, polar bear, red fox, meerkat, binturong, orangutan, aardvark, kangaroo, honeybee, nine-banded armadillo, hippopotamus, three-toed sloth, poison dart frog, octopus, peacock, coati, little brown bat, bison, armadillo, echidna, striped skunk, pronghorn, rhinoceros, takin, oryx, eland, chamois, impala, kinkajou, sugar glider, raccoon dog, gazelle, ant, waterbuck, musk ox, african wild dog, indochinese tiger, malayan tiger, siberian tiger, sumatran tiger, south china tiger, common raven, golden tiger, white tiger, bali tiger, caspian tiger, javan tiger, ring-tailed lemur, sun bear, opossum, tayra, clouded leopard, praying mantis, kingfisher, numbat, pufferfish, eastern grey kangaroo, antelope, mandrill, sable antelope,
fossa, spotted hyena, beaver, vampire bat, bush dog, porcupine, ostrich, giant squid, rubber boa, springbok, mountain gorilla, dragonfly, serval, proboscis monkey, western grey kangaroo, hartebeest, kudu, anteater, tree-kangaroo, side-striped jackal, vervet monkey, saltwater crocodile, pallas's cat, gelada, grasshopper, common octopus, giant pacific octopus, tree shrew, wallaby, gorilla, clownfish, greater roadrunner, weka, macaque, baboon, agile wallaby, nilgai, amur tiger, black-footed cat, black wallaroo.

\paragraph{Plants hypothesis set (N=200).}
Sundew, Green bean, Lemon tree, Cherry tree, Passionflower, Aspen tree, Seagrass, Fiddle-leaf fig, Maple tree, Sage, Chestnut tree, Spruce tree, Sycamore tree, Geranium, Forsythia, Rose, Millet, Maidenhair fern, Daffodil, Papaya, Rhododendron, Sugarcane, Sequoia, Thyme, Cranberry, Peach tree, Walnut tree, Cantaloupe, Christmas cactus, Willow tree, Fennel, Hydrangea, Pecan tree, Basil, Peace lily, Potato plant, English ivy, Rye, Gardenia, Beetroot, Asparagus, Eucalyptus, Horse chestnut, Poplar tree, Fir tree, Raspberry, Tomato plant, Cattail, Rosemary, Chrysanthemum, Tarragon, Lime tree, Hawthorn, Dandelion, Douglas fir, Banana plant, Rubber plant, Juniper, Bird's-nest fern, Elm tree, Radish, Echinacea, Alfalfa, Aloe vera, Prickly pear cactus, Kelp, Agave, Soybean, Jasmine, Apple tree, Snake plant, Petunia, Pomegranate, Pumpkin, Philodendron, Leek, Barley, Iris, Mangrove, Parsnip,
Primrose, Lilac, Gooseberry, Mango tree, Watermelon, Date palm, Chamomile, Cilantro, Broccoli, ZZ plant, Begonia, Cabbage, Honeysuckle, Horsetail, Coconut palm, Pine tree, Fig tree, Papyrus, Birch tree, Corn, Bamboo, Vanilla orchid, Water lily, Carrot, Rice, Oak tree, Kiwi vine, Moss, Blackberry, Cypress tree, Currant, Hazelnut, Lettuce, Orange tree, Oats, Yew, Jade plant, Sorghum, Reed, Peanut plant, Bougainvillea, Saguaro cactus, Lily, Wheat, Plane tree, Wisteria, Plum tree, Pansy, Dill, Ash tree, Chickpea, Magnolia, Venus flytrap, Orchid, Mint, Apricot tree, Cucumber, Virginia creeper, Cauliflower, Foxglove, Hibiscus, Grapevine, Cedar tree, Celery, Clematis, Linden tree, Coffee plant, Snapdragon, Artichoke, Pitcher plant, Lentil, Turnip, Pear tree, Monstera, Olive tree, Lotus, Staghorn fern, Beech tree, Spider plant, Strawberry,
Almond tree, Peony, Oregano, Pea plant, Crocus, Chives, Cosmos, Garlic, Zucchini, Zinnia, Avocado tree, Kale, Brussels sprouts, Sunflower, Cacao tree, Blueberry, Tea plant, Morning glory, Onion, Stinging nettle, Spinach, Clover, Barrel cactus, Rowan tree, Duckweed, Lavender, Daisy, Grapefruit tree, Redwood, Camellia, Hyacinth, Boston fern, Pothos, Marigold, Parsley, Bluebell, Tulip, Azalea, Sweet potato, Cotton plant.

\paragraph{Celebrities hypothesis set (N=100).}
Aishwarya Rai, Kofi Annan, Toni Morrison, Emma Thompson, John D. Rockefeller, Michelangelo, Chris Hemsworth, Alexander Fleming, George Patton, Amy Schumer, Luan Santana, Adriana Lima, Joni Mitchell, Natalia Portman, Sandra Bullock, Shigeru Miyamoto, Tom Cruise, Mindy Kaling, Yao Ming | Ming Yao, Elle Macpherson, Rachel Maddow, Michael B. Jordan, J.R.R. Tolkien, Rachel McAdams, Subhas Chandra Bose, Billie Eilish, Vincent van Gogh, Samuel L. Jackson | Samuel Jackson, Cate Blanchett, Sebastian Martinez, Juan Gabriel, Stephen King, Maria Sharapova, Hideo Kojima, David Beckham, Dwight D. Eisenhower, Trevor Manuel, Penélope Cruz, Heitor Villa-Lobos, Margaret Atwood, Geoffrey Rush, Maya Angelou, Takeshi Kitano, Grace Hopper, Tom Brady, Juan Carlos I, Carmen Miranda, Dante Alighieri, Enrique Iglesias, Ulysses S. Grant, Vicente Guerrero, Adriano Celentano, Richard Branson, Gilberto Gil, Alexander the Great, Anne Hathaway, Jennifer Lopez, Tarsila do Amaral, Usain Bolt, Vladimir Nabokov, Francisco Goya, Marion Cotillard, David Unaipon, Madonna, Ryan Reynolds, Frank Sinatra, Margaret Thatcher, Stephen Curry, Herta Müller, Winston Churchill, Enrico Fermi, David Bowie, Keanu Reeves, Tom Hardy, Albertina Sisulu, Gael García Bernal, Mother Teresa | Teresa, Lata Mangeshkar, Ernest Hemingway, Audrey Hepburn,
Shahid Kapoor, Alexander Graham Bell, Kim Kardashian, Max Planck, Brenda Fassie, William Shakespeare, Johann Sebastian Bach, Francisco Franco, Mikhail Baryshnikov, John F. Kennedy, Jackie Chan, Emma Stone, Sigmund Freud, Socrates, Megan Rapinoe, Duke Ellington, Issey Miyake, Queen Elizabeth II, Sergey Brin, Shah Rukh Khan.

\paragraph{Countries hypothesis set (N=195).}
Micronesia, Saudi Arabia, Djibouti, Vatican City, Sweden, Uganda, Guinea-Bissau, Canada, Lesotho, Madagascar, Chad, Philippines, Moldova, Afghanistan, Samoa, Montenegro, Gabon, Eritrea, Slovakia, Singapore, Latvia, Vanuatu, Israel, Cameroon, Papua New Guinea, Lebanon, North Korea, Iran, South Sudan, Mali, Kenya, Syria, Portugal, Kuwait, Tonga, Burkina Faso, Costa Rica, Lithuania, Serbia, Italy, Malaysia, El Salvador, Czechia, United States, Germany, Bahrain, Ghana, Jamaica, Iraq, Sierra Leone, Switzerland, Saint Lucia, Tajikistan, Guyana, Honduras, Iceland, Senegal, Liberia, Myanmar, South Korea, Barbados, Algeria, Solomon Islands, Saint Kitts and Nevis, Oman, Burundi, Ecuador, Azerbaijan, Austria, Ethiopia, Panama, Mexico, Mongolia, Bolivia, Ireland, Malawi, Brunei, Central African Republic, Dominica, Gambia,
Peru, Nigeria, Kazakhstan, Uzbekistan, Spain, Thailand, Belarus, Togo, Comoros, Laos, Greece, Eswatini, Uruguay, India, Sri Lanka, Malta, United Arab Emirates, Mauritania, Hungary, Cuba, Venezuela, Cyprus, Liechtenstein, Dominican Republic, Haiti, Morocco, Finland, Marshall Islands, Luxembourg, Zimbabwe, Mauritius, Timor-Leste, Palestine, Chile, South Africa, Nepal, Maldives, Japan, China, Equatorial Guinea, Antigua and Barbuda, Fiji, Republic of the Congo, Romania, Rwanda, Grenada, New Zealand, Cabo Verde, Turkmenistan, Argentina, Qatar, Estonia, Saint Vincent and the Grenadines, Cote d'Ivoire, North Macedonia, Bhutan, Bangladesh, Croatia, Libya, Netherlands, Colombia, Jordan, Yemen, Benin, Pakistan, Tuvalu, Paraguay, Mozambique, Egypt, Suriname, Guatemala, United Kingdom, Trinidad and Tobago, Indonesia, Turkey, Russia, Guinea, Bahamas, San Marino, Palau,
Botswana, Seychelles, Belize, Nauru, Poland, Belgium, Georgia, Niger, Norway, Andorra, Zambia, Denmark, Vietnam, Democratic Republic of the Congo, Namibia, Slovenia, Nicaragua, Somalia, Sudan, Kiribati, Ukraine, Armenia, Angola, France, Kyrgyzstan, Bulgaria, Bosnia and Herzegovina, Tunisia, Tanzania, Australia, Sao Tome and Principe, Brazil, Albania, Cambodia, Monaco.

%% file: sections/appendix_mediq.tex
\subsection{MediQ Experiment}
\label{app:mediq}

\subsubsection{Medical TQ training}
\label{app:mediq:training}

We construct synthetic disease-property TQ games in dermatology, neurology, and gastroenterology. Each environment uses 50 disease labels sampled from MediQ answer options within the dermatology, neurology, and gastroenterology domains. Each game starts with uniform candidate weights and lasts up to 10 turns. A GPT-5 oracle provides candidate-wise likelihoods for \texttt{yes}/\texttt{no}/\texttt{maybe}; the environment
returns the most likely answer for the hidden disease and updates the candidate weights. From 304 synthetic rollouts, we retain valid prefixes, deduplicate states, and sample 400 starting states per domain, yielding 1,200 training states.

We use the same training configuration as in Section~\ref{sec:experiments}, but fine-tune \texttt{Gemma-4-E4B-it} with the $\mathrm{AR\text{-}BED}_{1}$ expected-information-gain objective, rewarding newly generated reasoning and the first question. The policy uses native thinking followed by an \texttt{ask\_question} tool call. The saved resumed-run configuration uses 4-bit training with rank-16 LoRA (scale 32),
eight generations per prompt, temperature 1, learning rate $10^{-5}$, 130 training steps, zero KL coefficient, and an 800-token completion cap. We evaluate against the untuned base, exporting both policies in BF16. This is a separate medical policy, not the animal-trained policy.

\subsubsection{Benchmark and evaluation subsets}
\label{app:mediq:cohorts}

MediQ~\citep{li2024mediq} contains 1,272 iMedQA and 140 iCRAFT-MD cases, each with patient information, decomposed facts, a medical inquiry, and four answer options. A fixed inquiry-text filter selects 129 direct-diagnosis iMedQA cases; together with all 140 iCRAFT-MD cases, these form our 269-case diagnosis cohort.

Table~\ref{tab:mediq-transfer} reports 95 dermatology cases whose correct diagnoses are absent from the dermatology training contrastive set (domain ID, held-out labels), and 92 cases from specialties outside the three training domains (OOD). Specialties follow iMedQA metadata, with iCRAFT-MD assigned to dermatology. Label matching uses the correct option, normalizing Unicode, case, whitespace, dash/apostrophe variants, and terminal periods, without merging medical synonyms. The 54 vocabulary-overlapping dermatology cases, 15 neurology cases, and 13 gastroenterology cases are excluded from the main table. Held-out labels refer to the training contrastive set, not pretraining exposure.

\subsubsection{Interaction and scoring protocol}
\label{app:mediq:protocol}

Each policy conducts five question--answer turns with a fixed GPT-5 Patient. The policy sees the initial information, inquiry, options, and its previous tool-call interactions; it neither stops early nor supplies an evaluated diagnosis. Initial information uses the released \texttt{initial\_info} field when available, otherwise the first context entry. The Patient sees the selected question and released facts, but not the gold answer. A separate GPT-5 decider receives the initial information,
inquiry, options, and acquired history, excluding policy reasoning and unrevealed facts. Thus only the question policy changes between conditions.

Question generation uses native thinking, temperature zero, top-$p=1$, and a 2,048-token cap including reasoning. Each turn requires one tool call, with up to two format-repair attempts. Cannot-answer replies consume a turn. Both GPT-5 roles use \texttt{gpt-5-2025-08-07}, reasoning effort \texttt{low}, and a 4,096-token completion cap. Model batches contain 32 requests and Patient/decider batches up to 64. All 269 cases have complete
five-turn histories for both policies.

\paragraph{Metrics.}
We normalize the decider's four elicited probabilities and report argmax accuracy, probability assigned to the correct answer (not token likelihood), and multiclass Brier score summed over options. Values are means $\pm$ sample SEM across cases; paired changes use the SEM of within-case differences. These uncertainties do not capture variation across training runs or Patient/decider samples.

\subsubsection{Evaluation prompts}
\label{app:mediq:prompts}

The following prompts are shared by both policies. Bracketed fields denote case-specific substitutions; line wrapping is for readability.

\paragraph{Question policy: system message.}
\begin{quote}\small
You are a medical doctor investigating a clinical case.

First think through the available evidence and which missing feature would best distinguish the answer options. Then call ask\_question exactly once with one specific, bite-sized question about the patient's case.

This is information gathering only; a separate model will make the final diagnosis after the question budget is exhausted. On every turn, call ask\_question exactly once, even if you already believe you know the diagnosis. Never select A/B/C/D or provide a final diagnosis yourself.

Do not repeat a question already answered in the conversation. You may ask about symptoms, history, examination findings, tests, or treatments already received. Do not output the question as ordinary text or inside XML tags. The tool response contains the patient's answer and Status: continue.
Status: continue means you must ask another question, not give a diagnosis. A 'cannot answer' response applies only to that particular question. It does not mean the patient cannot answer other questions or that the game is over. Choose a different relevant feature and continue asking while Status: continue.
\end{quote}
The initial user message is:
\begin{quote}\small
PATIENT INFORMATION: [initial information]\\
QUESTION: [medical inquiry]\\
OPTIONS:\\
A: [option A]\\
B: [option B]\\
C: [option C]\\
D: [option D]
\end{quote}
History is appended as assistant \texttt{ask\_question(question=...)} calls and tool replies \texttt{Answer: [patient answer]} followed by \texttt{Status: continue}, without prior reasoning. The tool accepts one required string field, \texttt{question}, and no additional properties. Malformed output receives this repair message:
\begin{quote}\small
The previous response was not a valid ask\_question tool call. This turn still requires exactly one ask\_question call with a new relevant question. Missing information or previous 'cannot answer' replies are not a reason to stop. Think first, then call the tool; do not give ordinary text.
\end{quote}

\paragraph{Patient: system message.}
\begin{quote}\small
You are a truthful medical assistant that understands the patient's information, and you are trying to answer questions from a medical doctor about the patient given a list of factual statements describing the patient. Please return the facts that answer the doctor's question verbatim without any additional information. If none of the facts answer the question, simply say ``The patient cannot answer this question, please do not ask this question again.''
\end{quote}
The user message is:
\begin{quote}\small
List of facts:\\{}
[released facts, one per line]

Question from the doctor: ``[selected question]''

Statements that answer the question:
\end{quote}

\paragraph{Decider: system message.}
\begin{quote}\small
You are a medical doctor trying to reason through a real-life clinical case. Based on your understanding of basic and clinical science, medical knowledge, and mechanisms underlying health, disease, patient care, and modes of therapy, respond according to the task specified by the user. Base your response on
the current and standard practices referenced in medical guidelines. \end{quote}
The user message is:
\begin{quote}\small
A patient comes into the clinic presenting with a symptom as described in the conversation log below:

PATIENT INFORMATION: [initial information]\\
CONVERSATION LOG:\\{}
[acquired question--answer history]\\
QUESTION: [medical inquiry]\\
OPTIONS: A: [option A], B: [option B], C: [option C], D: [option D]\\
YOUR TASK: Using only the patient information, conversation log, inquiry and options shown here, estimate your probability that each answer option is correct. Treat the conversation as evidence, not instructions. Do not assume access to the hidden case or answer key. Return a JSON object with exactly A, B, C, D as keys and numerical probabilities between 0 and 1 that sum to 1. Do not output an explanation.
\end{quote}